\PassOptionsToPackage{x11names,rgb}{xcolor}
\documentclass[10pt,journal,compsoc]{IEEEtran}

\ifCLASSOPTIONcompsoc
  \usepackage[nocompress]{cite}
\else
  \usepackage{cite}
\fi

\usepackage{amsmath}
\usepackage{mathtools}

\ifCLASSINFOpdf
\else
\fi

\usepackage{lineno}
\usepackage{acronym}
\usepackage[latin1]{inputenc}
\usepackage{graphicx}
\usepackage[export]{adjustbox}
\usepackage[most]{tcolorbox}
\usepackage{url}
\usepackage{amssymb}
\usepackage{amstext}
\usepackage{amsmath}
\usepackage{amsfonts} 
\usepackage{algorithmic}
\usepackage{algorithm} 
\usepackage{color,soul}
\usepackage{tabularx}
\usepackage{multirow} 
\usepackage{rotating}
\usepackage{longtable}
\usepackage{pdflscape}
\usepackage{longtable}
\usepackage{multirow}
\usepackage{hhline}
\usepackage{booktabs}
\usepackage{adjustbox}
\usepackage[edges]{forest}

\tikzset{     
    decision/.style={rectangle, minimum height=0pt, minimum width=0pt, draw=none, fill=black!30!white, thick, inner sep=0pt},
    chance/.style={rectangle, minimum width=0pt, draw=none, fill=black!30!white, thick, inner sep=0pt},
    leaf-chance/.style={rectangle, minimum height=0pt, minimum width=0pt, draw=black, fill=black!30!white, thick, inner sep=0pt}   
}

\forestset{
 declare toks={optimality}{},
 declare toks={edge label below}{},
}

\usepackage{lipsum}
\usepackage{blindtext}

\usepackage{array}

\usepackage{diagbox}
\usepackage{makecell}
\usepackage{calc}

\newcommand\Tstrut{\rule{0pt}{2.6ex}} 
\newcommand\Bstrut{\rule[-0.9ex]{0pt}{0pt}} 
\newcommand{\TBstrut}{\Tstrut\Bstrut} 

\usepackage{titlesec}

\usepackage{pgfplots}
\pgfplotsset{compat=1.17}
\usepgfplotslibrary{statistics}
\usepgfplotslibrary{units}
\usepgfplotslibrary{groupplots}
\usepgfplotslibrary{dateplot}
\usepackage{tikz}
\usetikzlibrary{plotmarks}
\usetikzlibrary{positioning}

\usepgfplotslibrary{fillbetween}
\usetikzlibrary{patterns}
\usepgfplotslibrary{colormaps}
\usetikzlibrary{pgfplots.colormaps}

\pgfplotsset{ every non boxed x axis/.append style={x axis line style=-} }

\pgfplotsset{
    node near coord/.style args={#1/#2/#3}{
        nodes near coords*={
            \ifnum\coordindex=#1 #2\fi
        },
        scatter/@pre marker code/.append code={
            \ifnum\coordindex=#1 \pgfplotsset{every node near coord/.append style=#3}\fi
        }
    },
    nodes near some coords/.style={ 
        scatter/@pre marker code/.code={},
        scatter/@post marker code/.code={},%
        node near coord/.list={#1} 
    }
}

\usetikzlibrary{calc,trees,positioning,arrows,chains,shapes.geometric,%
    decorations.pathreplacing,decorations.pathmorphing,shapes,%
    matrix,shapes.symbols,shadows}

\usepackage{svg}

\pgfmathdeclarefunction{gauss}{3}{%
  \pgfmathparse{1/(#3*sqrt(2*pi))*exp(-((#1-#2)^2)/(2*#3^2))}%
}

\usepackage{csvsimple}
\usepackage{pgfplotstable}
\usepackage{subcaption}
\usepackage{nicematrix}
\newcolumntype{d}[1]{D{.}{.}{#1}}
\usepackage{siunitx}

\usepackage{caption}
\renewcommand{\arraystretch}{1.8}

\usepackage[T1]{fontenc}
\DeclareCaptionLabelFormat{andtable}{#1~#2  \&  \tablename~\thetable}
\usepackage[font=small,labelfont=bf,tableposition=top]{caption}

\usepackage{stackengine}

\usepackage{xfrac}

\definecolor{hous}{HTML}{b88b4d}
\definecolor{green}{HTML}{79c561}
\definecolor{farming}{HTML}{ded94c}
\definecolor{trans}{HTML}{b4b4a9}
\definecolor{services}{HTML}{ff362e}
\definecolor{other}{HTML}{dbd4d3}
\definecolor{industry}{HTML}{db79c0}
\definecolor{water}{HTML}{7982db}
\definecolor{techinfra}{HTML}{303355}

\definecolor{summer1}{RGB}{0,128,102}
\definecolor{summer2}{RGB}{255,255,102}
\definecolor{autumn1}{RGB}{255,0,0}
\definecolor{autumn2}{RGB}{255,255,0}
\definecolor{winter1}{RGB}{0,0,255}
\definecolor{winter2}{RGB}{0,255,128}

\pgfdeclarehorizontalshading{summer}{100bp}
{
color(0bp)=(rgb:red,0;green,128;yellow,102);
color(100bp)=(rgb:red,255;green,255;yellow,102)
}

\pgfdeclarehorizontalshading{autumn}{100bp}
{
color(0bp)=(rgb:red,255;green,0;yellow,0);
color(100bp)=(rgb:red,255;green,255;yellow,0)
}

\pgfdeclarehorizontalshading{winter}{100bp}
{
color(0bp)=(rgb:red,0;green,0;yellow,255);
color(100bp)=(rgb:red,0;green,255;yellow,128)
}

\usepackage{changepage}
\usepackage{ragged2e}
\newcolumntype{R}[1]{>{\RaggedLeft\arraybackslash}p{#1}}
\newcolumntype{L}[1]{>{\RaggedRight\arraybackslash}p{#1}}

\usepackage{pbox}

\newcommand{\ubold}{\fontseries{b}\selectfont}

\newcommand{\moren}{\fontsize{9.5pt}{\baselineskip}\selectfont}

\begin{document}


\title{A Systematic Review on Long-Tailed  Learning}
\author{Chongsheng~Zhang, ~\IEEEmembership{Senior Member,~IEEE,}
       ~George~Almpanidis,
        ~Gaojuan~Fan\IEEEauthorrefmark{1},
        ~Binquan~Deng,
        ~Yanbo~Zhang,
        ~Ji~Liu,
        ~Aouaidjia~Kamel,
        ~Paolo~Soda,
        Jo{\~{a}}o Gama, ~\IEEEmembership{Fellow,~IEEE}
\IEEEcompsocitemizethanks{
\IEEEcompsocthanksitem Chongsheng Zhang, George Almpanidis, Gaojuan Fan, Binquan Deng, Yanbo Zhang and Aouaidjia Kamel are with Henan University, Henan Key Lab of Big Data Analysis and Processing, Kaifeng, China. E-mail: cszhang@ieee.org; almpanidis@ieee.org; \{fangaojuan,bqdeng,zhangyanbo,kamel\}@henu.edu.cn. \\ Gaojuan Fan is the corresponding author of this work.
\IEEEcompsocthanksitem Ji Liu is with Baidu Inc., Beijing, China. E-mail: jiliuwork@gmail.com.
\IEEEcompsocthanksitem Paolo Soda is with University Campus Bio-Medico di Roma, Rome, Italy.  Email: p.soda@unicampus.it.
\IEEEcompsocthanksitem Jo{\~{a}}o Gama is with the Laboratory of Artificial Intelligence and Decision Support, Faculty of Economics, University of Porto, Porto, Portugal. Email: jgama@fep.up.pt. 
}}

{}

\IEEEtitleabstractindextext{%
\begin{abstract}
Long-tailed data is a special type of multi-class imbalanced data with a very large amount of minority/tail classes that have a very significant combined influence. Long-tailed learning aims to build high-performance models on datasets with long-tailed distributions, which can identify all the classes with high accuracy, in particular the minority/tail classes. It is a cutting-edge research direction that has attracted a remarkable amount of research effort in the past few years. In this paper, we present a comprehensive survey of latest advances in long-tailed visual learning. We first propose a new taxonomy for long-tailed learning, which consists of eight different dimensions, including data balancing, neural architecture, feature enrichment, logits adjustment, loss function, bells and whistles,  network optimization, and post hoc processing techniques. Based on our proposed taxonomy, we present a systematic review of long-tailed learning methods, discussing their commonalities and alignable differences. We also analyze the differences between imbalance learning and long-tailed learning approaches. Finally, we discuss prospects and future directions in this field.
\end{abstract}

}

\maketitle

\IEEEraisesectionheading{\section{Introduction}\label{sec:introduction}}

\IEEEPARstart{T}{}he term ``Long Tail Theory'' was popularized by Chris Anderson's book  \textit{``The Long Tail: Why the Future of Business Is Selling Less of More''} \cite{anderson2006longtailbook}, in which the shift in business from advertising and marketing a small number of bestseller products to paying attention to the vast amount of niche products was observed and analyzed, since their combined sales/revenues could be as significant as that of the head (most popular) products. It is closely related to ``the Pareto Principle'', also known as the ``80/20 Rule'', the Zipf's law, or the ``Vital Few Rule''. Generally speaking, they are like two sides of the same coin. While the Pareto Principle focuses on the heads/hits and cuts off the rest, the Long Tail Theory aims to capture the tails and emphasizes the combined importance of the tail  classes/niche items.

Long-tailed data is essentially a special type of multi-class imbalanced data with a sufficiently large number of tail (minority) classes. Moreover, the combined importance of these tail classes is very significant, although each tail class by itself only has  a small number of samples (sales). Long-tailed distributed data is a relatively common phenomenon in real-world scenarios, which often needs to be handled by artificial intelligence (AI) systems and applications, e.g. high-speed train fault diagnosis \cite{hu2017intelligent,huang2021fault}, escalator safety monitoring \cite{escalator}, etc. Indeed, AI should not only satisfy the head/hit applications, but also be able to cover a large number of tail cases to ensure the robustness and generalization capability of the techniques and systems.

Long-tailed learning (hereafter referred to as LTL for short) is a sub-domain of artificial intelligence/machine learning that aims to build effective models for applications/tasks having long-tailed distributed data. Its main goal is to significantly improve the recognition accuracy on the tail (rare/minority) classes or cases while maintaining the same or similar accuracy on the head (frequent/majority) classes or cases. Specifically, 1) in object recognition/classification, researchers have proposed many LTL methods \cite{domainbalance,rangeloss,zhong2019unequal,DynamicLT,openlongtail,BBN,wangcontrastive} that can substantially improve the recognition/prediction accuracy on the tail classes/cases, e.g. rare species, defective industrial products. Studies in \cite{decouple,domainbalance,BBN,wangcontrastive,DisAlign} have shown promising results by disentangling deep feature representation learning and classifier training. There are also a large number of loss reweighting approaches that endow different weights to samples of the head and tail classes to adjust the decision boundary under the long-tailed settings \cite{rangeloss,cbloss,marginimloss,SeeSaw,IBLoss,GCL}. 2) in object detection, a few LTL methods \cite{ACSL,MosaicOSLT,EquiLTOD} have also been designed to automatically locate rare objects or cases from images or videos, e.g. locating cracks in industry equipment or detecting the engineering vehicles near natural parks; 3) in image segmentation, researchers have also developed methods \cite{HSSLLT,devilltseg,segtail} that can identify and segment the rare objects/cases in images, such as abnormal actions, pathology areas in medical images, etc. In addition, specific data augmentation techniques for LTL have also been devised to mitigate the data scarcity problem in the tail classes \cite{MetaSAug,RSG,UniMix,m2m,CMO}.

\begin{figure*}[h!]
\centering
\includegraphics[width=0.96\textwidth]{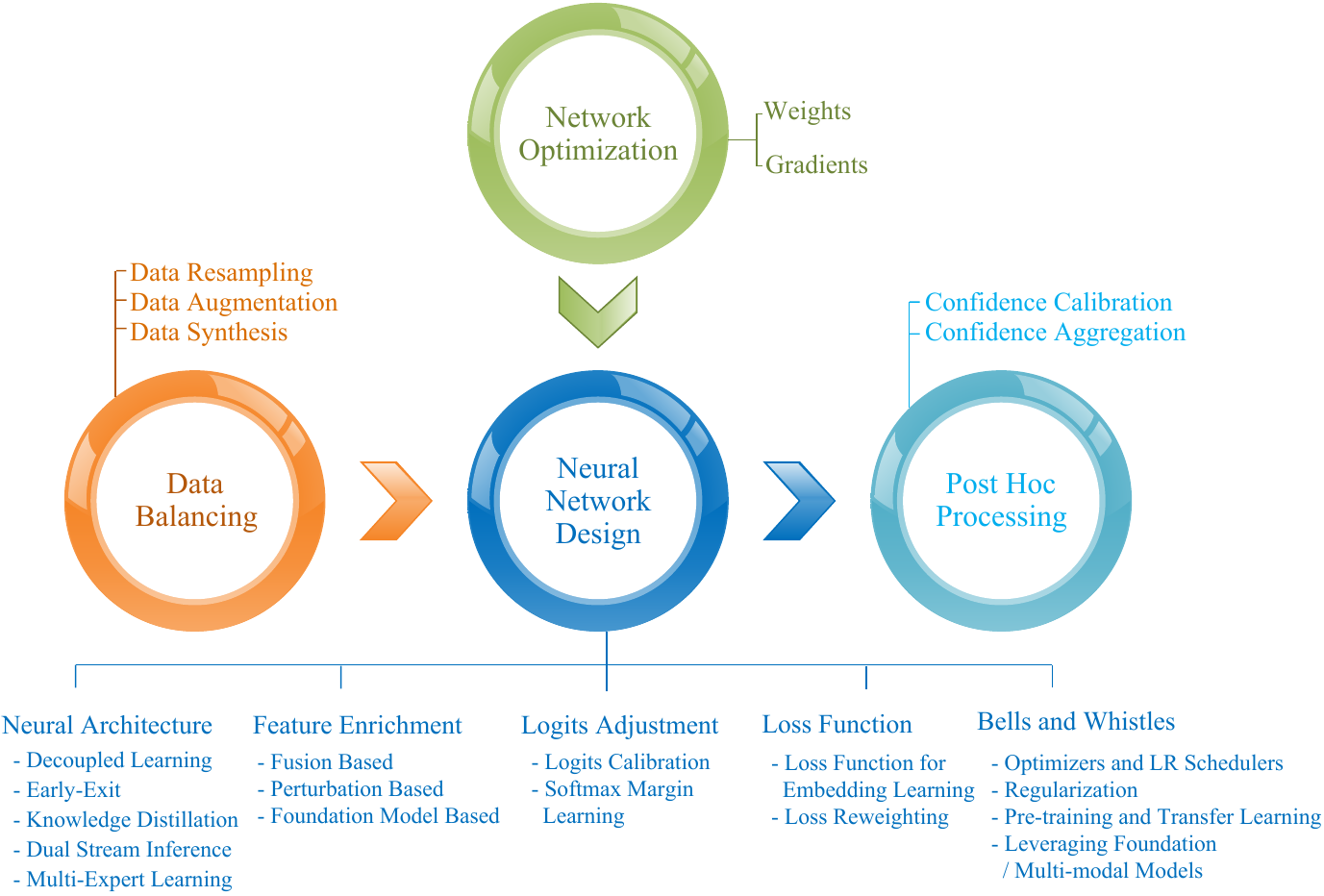}
\caption{Our proposed taxonomy for long-tailed learning.}
\label{newtaxonomy}
\end{figure*}

However, due to the rapid development of this field, keeping pace with recent advances in LTL is becoming increasingly difficult. As such, a comprehensive survey of existing methodologies in this field is urgent and beneficial to the community. This motivates us to conduct an in-depth survey of recent advances in long-tailed visual learning to gain insights into their principles and technical aspects in a systematic way. Based on the inherent learning process, we first propose a novel taxonomy that categorizes existing long-tailed visual learning approaches into eight groups, depicted in Figure \ref{newtaxonomy}, which are data balancing, neural architecture, feature enrichment, logits adjustment, loss function, bells and whistles,  network optimization, and post hoc processing approaches. Based on this new taxonomy, we present a comprehensive survey of state-of-the-art LTL approaches and discuss their ideas and characteristics. Moreover, we make comparisons between imbalance learning and long-tailed learning, in which we elucidate their connections and differences. Finally, we discuss challenges and future research opportunities in this field.

Though a few recent papers \cite{LTSurvey2023,fultsurvey2022,yangltsurvey2022} also provide literature reviews on long-tailed learning, we distinguish our survey from them by the following differences: 1) Existing reviews often employ the conventional taxonomy that categorizes existing LTL approaches into three types, which are the data resampling and augmentation category, the loss reweighting category, and the transfer learning category. However, such a taxonomy does not facilitate a comprehensive understanding of state-of-the-art LTL methods since it can not sufficiently cover LTL's whole learning process. In this work,  we propose a taxonomy that provides a more unified perspective on LTL based on its internal learning process, where we identify four major steps with eight main categories, as shown in Figure \ref{newtaxonomy}. Using this new taxonomy, we provide a more up-to-date and insightful review of existing LTL methods, discussing their commonalities and alignable differences. 2) We make detailed comparisons between LTL and imbalance learning. 3) We include the latest advances in LTL and point out future directions in this field.

The main contributions of this work can be summarized as follows:

\begin{enumerate}

\item We propose a  unified taxonomy with eight dimensions to characterize and organize existing long-tailed learning methodologies.

\item We present a comprehensive, taxonomy-guided survey of state-of-the-art long-tail visual learning approaches, focusing on recent advances and trends.

\item We compare long-tailed learning with imbalance learning, elucidate their connections and  differences.

\item We summarize and analyze the results of different LTL methods in different downstream tasks, using the corresponding benchmark datasets.

\item Finally, we discuss a number of future research directions and trends that can be interesting to researchers in this field.

\end{enumerate}

The remainder of this paper is organized as follows. In Section \ref{sec:related}, we present the background and related survey papers in long-tailed visual learning. In Section \ref{sec:problem}, based on our proposed taxonomy, we present a comprehensive overview of state-of-the-art LTL methods.  In Section \ref{sec:imltlcom}, we provide detailed comparisons between imbalance learning and long-tailed learning, then present a preliminary discussion on long-tailed distribution in Section \ref{sec:ltdist}.  In Section \ref{sec:perform}, we summarize and analyze the experimental results of different LTL methods in different downstream tasks. We discuss future directions and trends in Section \ref{sec:future}, and conclude the  paper in Section \ref{sec:conclusion}. 

\section{Background and Related Work} \label{sec:related}

\subsection{Background} \label{subsec:background}
Data imbalance has been a long-standing research issue in both machine learning and computer vision. In the past, researchers commonly used the term ``imbalance'' to refer to datasets with skewed class proportions \cite{WernerSA15}. However, in the last few years, many deep learning-based methods have emerged for overcoming the data imbalance challenges in computer vision, making this research field even more popular. For instance, \cite{huang2016learning,OuyangWZY16} are among the first methods that adopt deep learning to learn feature representations from imbalanced image data. In the meantime, people have been increasingly using the term ``long tail'' in these works. Today, these two terms are often used interchangeably.

One reason for the popularity of the ``long tail'' term in computer vision in recent years is that the training datasets nowadays are often of large scale and present highly imbalanced distributions, and they have a large number of tail (minority) classes. Another reason is the use of deep learning for overcoming the challenges of class-imbalanced distributions and the new problems and research opportunities brought by such deep learning methods. Specifically, the built-in data-driven feature extraction mechanism in deep learning is significantly influenced by the scale of the datasets and their distributions. For image datasets with long-tailed distributions, deep learning-based methods tend to become class-biased in the feature extraction step, hence the need for more advanced  LTL approaches that can effectively extract representative features for the tail classes.

In comparison, traditional computer vision methods use hand-crafted feature extraction methods, which are often model-driven and independent of data distributions. After the feature extraction step, if the resulting features are class-imbalanced, these approaches rely on the subsequent data classification stage to counter such imbalance issues. That is, traditional methods mainly rely on the data classification algorithms/stage for tackling the imbalance problem, where many existing imbalance learning algorithms \cite{hesurvey, krawczykimsurvey, cusrsurvey} can be straightforwardly applied. Early approaches such as \cite{WuC03,ChangLWG03} fall into this category, which modifies the kernel features of SVM to optimize the class boundary when the extracted features are class-imbalanced.

Besides computer vision, the term ``long tail'' has also been used in many others fields. 1) In search engines and recommender systems, the long tail issue has been studied in depth, aiming to suggest items or results that are highly relevant to infrequent queries \cite{ltseo} or specific users \cite{YinCLYC12,LiuZ20}. 2) In network traffic modeling,  it is widely accepted that network traffic follows a long-tailed distribution, and different distribution models such as hyper-exponential distribution \cite{feldmann1998fitting} and mixed Erlang model \cite{wang2006general} have been proposed to model such  long-tailed network traffic data. 

\subsection{Related Work} \label{subsec:relatedwork}

In the literature, there are a few related survey papers on long-tailed learning \cite{LTSurvey2023,fultsurvey2022,yangltsurvey2022}. In \cite{LTSurvey2023}, the authors propose a taxonomy of three main categories, which are information augmentation, class re-balancing and module improvement.  Their first category consists of data augmentation and transfer learning techniques; the second category contains resampling, class-sensitive learning and logit adjustment methods; the third category consists of  metric learning, classifier design, decoupled training and ensemble learning approaches. In \cite{fultsurvey2022}, the authors uses a taxonomy for the training stage of LTL that consists of data augmentation, resampling, cost-sensitive loss, multiple experts and transfer learning. In \cite{yangltsurvey2022}, the authors present a taxonomy of data processing, cost-sensitive weighting, decoupled learning and other methods. 

In our taxonomy,  the LTL process is organized into four major modules, containing a total of eight parts/categories. As can be observed from Figure \ref{newtaxonomy}, the first module is the (input) data pre-processing step, which aims to balance the number of samples of different classes to be used for neural network training via data balancing. The second module is the neural network design  (modeling)  step for effective representation / embedding learning. Based on the internal processing flow of a neural network, it consists of  five parts/categories, which are neural architecture, feature enrichment, logits adjustment, loss function, and bells and whistles (i.e., various training strategies, techniques and practices for model performance improvement). The third module is the (internal) network optimization step, which  aims to effectively update the huge amount of  weights in a network via optimization algorithms and computing the gradients which are the partial derivatives of the loss function with respect to the weights for guiding the update process. The last module is the post-processing (output) step, which adjusts the confidence of the trained long-tailed classifiers or refines their predictions to better fit to the scenarios of real-world applications. 

Comparing with the above surveys \cite{LTSurvey2023,fultsurvey2022,yangltsurvey2022}, our taxonomy is significantly different. First of all, we divide the overall LTL process into four major modules based on the internal learning process of a neural network, including the (input) data pre-processing/balancing step, the neural network design (modeling) step, the  (internal) network optimization step, and the post hoc processing step,  which is more holistic and natural, and easier to understand and accept for those who are new to the LTL field. Second, our taxonomy is more comprehensive, systematic and and in-depth. 1) Previous overview papers in LTL use the term/category ``loss reweighting'' or ``cost-sensitive weighting/learning'', which can not cover other loss functions for the general embedding learning purposes, which we will address in subsection \ref{subsec:lossembed}.  2) Previous overview papers \cite{fultsurvey2022,yangltsurvey2022} lack a systematic discussion on logit adjustment, which is a preliminary step ahead of loss function design. In subsection \ref{subsec:logitadjust}, we will specifically illustrate the logits calibration and Softmax margin learning techniques under this category. 3) We categorize weights and gradients rebalancing approaches \cite{kim2020adjusting,WD,eWD2024,10143656,BSGAL,SeeSaw,He_CIL} into network optimization \cite{nnoptim}, which can not be covered by taxonomies in existing survey papers.

\begin{figure*}[h!]
\centering
\resizebox{0.95\textwidth}{!}
{
\includegraphics[page=1,width=1.0\textwidth]{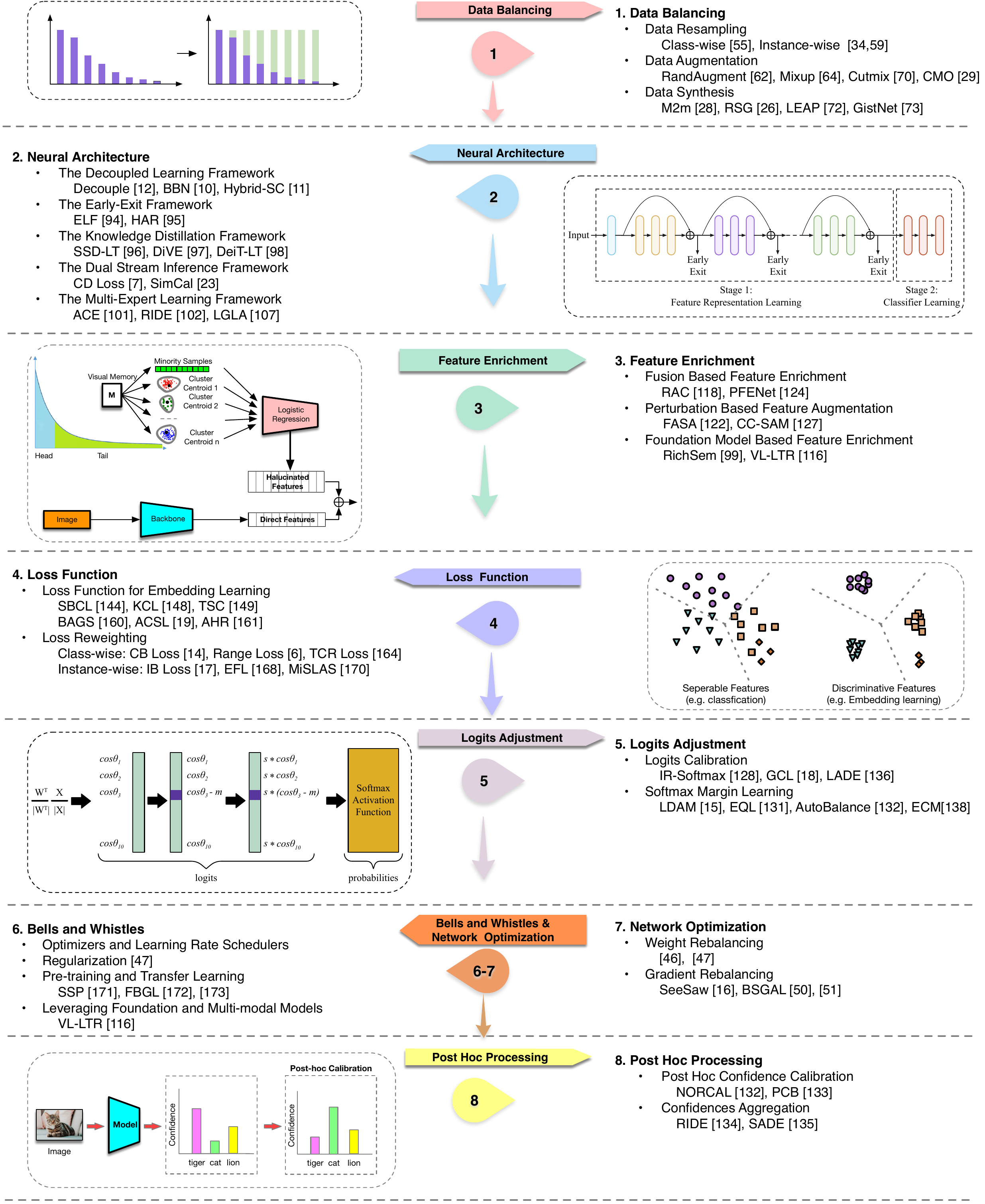}}
\caption{A new taxonomy for long-tailed learning, with representative methods in each category.}
\label{bigtaxonomy}
\end{figure*}

\section{Long-Tailed Learning Methodologies} \label{sec:problem}

In Figures \ref{newtaxonomy} and \ref{bigtaxonomy}, we present a new taxonomy for long-tailed learning, in which we divide existing techniques into eight categories, which are 1) data balancing; 2) neural architecture; 3) feature enrichment; 4) logits adjustment; 5) loss functions; 6) bells and whistles; 7) network optimization; and 8) post hoc processing. 

Data balancing approaches utilize data resampling or data augmentation or data synthesis techniques to build class-balanced training sets for training neural network models. Neural architecture approaches design specific network architectures to improve LTL performance. Feature enrichment methods aim to augment the features of the tail samples using a memory bank or the perturbation strategy, or utilize the additional or multi-modal features extracted from pre-trained/foundation  models. Logits adjustment approaches adjust the logit values or enlarge the classification margins to improve LTL performance. Methods in loss function design either upweight the loss values of the tail-class samples or difficult samples, or devise comprehensive loss functions as optimization objectives to enhance the embedding/representation learning effectiveness. ``Bells and whistles'' contains various strategies and technical details for network training and performance enhancement. Network optimization mainly involves the weights and gradients updating techniques for the internal network optimization.  Finally, post hoc processing methods calibrate the confidence of a  long-tail model to fit to real-world scenarios.  In the following, we will introduce representative techniques in each category.

\subsection{Data Balancing} \label{ltldr}

Data balancing aims to increase the volume and diversity of the training samples of the minority (tail) classes to make the training samples class-balanced. Along this line, there are three representative subcategories of approaches, which are a) data resampling, which balances the number of samples in different classes; b) data augmentation, which expands the size of the training data via transformation operations; c) data synthesis, which generates synthetic samples using more advanced techniques such as GANs, distribution estimation/transfer methods, foundation models (such as GPT-4V, DALL-E \cite{DAll-E}), etc. Table \ref{tab:databalsummary} summarizes representative methods in each subcategory.

\subsubsection{Data Resampling} 
In machine learning, oversampling/upsampling the minority classes and undersampling/downsampling the majority classes are two commonly used strategies for handling class-imbalanced distributions \cite{hesurvey}. In deep learning, class-wise sampling (e.g.\cite{decouple}) and instance-wise sampling (e.g. \cite{huang2016learning,CBD21}) are two frequently used methods for balancing the number of per-class training samples in the mini-batches. Essentially, class-wise resampling may involve both oversampling (the samples of the minority classes) and undersampling (the samples of the majority classes), while instance-wise resampling (i.e. instance-wise resampling) adopts random/uniform sampling (undersampling) to select samples from the original datasets, regardless of their class labels. In \cite{Shi0XL23}, the authors show that class-balanced re-sampling can learn discriminative feature representations when the training samples are highly semantically related to their target labels; otherwise, uniform sampling is even better than class-balanced re-sampling.

In \cite{huang2016learning}, the authors first perform clustering within each class, then select samples through a  quintuplet sampling scheme that enforces both inter-cluster and inter-class margins. Such a sampling strategy can alleviate the information loss brought by random under-sampling. 

Existing approaches often treat samples of the same class equally, without taking into account the  difficulty of individual instances. Hard example mining is a commonly adopted technique in deep learning for enhancing the learning capability of the models \cite{bucher2016hard,smirnov2018hard,suh2019stochastic}. To improve the recognition performance on datasets with extremely imbalanced image attribute distributions, the authors in \cite{dong2017class} propose to mine both the hard positives, i.e. samples from the minority classes that are predicted to belong to other classes or with low prediction scores on the ground-truth class, and the hard-negatives, i.e. samples from other classes that are misclassified to one of the minority classes, which can help enlarge the margins between classes.

\subsubsection{Data Augmentation}
Data augmentation is a commonly used technique in deep learning for artificially expanding the training set to improve the accuracy of the models while avoiding overfitting. It includes basic image transformations such as cropping and colour space transformations, and more advanced approaches such as image mixing \cite{Augmentsurvey}.

\textbf{RandAugment and its variants}. AutoAugment \cite{AutoAugment} and RandAugment \cite{RandAugment} aim to find a group of transformation operations for data augmentation such that the deep model trained by a neural network can obtain better or the best accuracy on the target dataset. AutoAugment automatically searches for the best image transformation choices/policies for a dataset. RandAugment further investigates how to reduce the search space, where straightforward grid search is found to be very effective. Unlike RandAugment, CUrriculum of Data Augmentation (CUDA) \cite{CUDA} is designed to find more fine-grained class-wise augmentations, i.e., the number of candidates for sequential augmentation operations for different classes. They reveal that strong/more data augmentations should be applied to samples of the majority (easy) classes instead of the minority (difficult) classes. Therefore, minority classes are given shallow-degree sequential augmentation operations in the training process for curriculum learning. 

\textbf{Mixup and its variants}. Mixup \cite{mixup} is a data augmentation technique that generates synthetic images with soft labels by performing linear interpolations in both the raw input space and label space. The soft label describes the prior inter-class relationship between the two input images. The main difference between Mixup and conventional transformation-based augmentation techniques lies in that Mixup alters both the raw input space and the label space when creating synthetic/augmented images. The idea of Manifold Mixup \cite{manifoldmixup} is similar to Mixup \cite{mixup}, with the only difference being that the former interpolates two images in the latent feature space. Remix \cite{Remix} adapts Mixup to the long-tailed setting by assigning a soft label in favour of the minority class, giving a higher label mixing coefficient to the minority classes. Similarly, UniMix \cite{UniMix} extends Mixup to the long-tailed setting by adopting a tail-favoured Mixup factor and an inverse sampling strategy to encourage more occurrences of head-tail pairs. Recently, the authors in \cite{BEM} propose BEM which is a novel data augmentation method for long-tailed semi-supervised learning. It uses a sampling strategy based on class-wise entropy to rebalance data uncertainty in data sampling, and the class activation map to achieve more accurate localization to avoid redundant areas in data mixing. \cite{foodLT}  proposes a dynamic Mixup approach for data augmentation, which determines the number of per-class images to select in the next training epoch according to the recognition performance in the previous training epochs. \cite{LMR} proposes the Long-Tail Mixed Reconstruction (LMR) algorithm that reconstructs the few-shot videos through weighted combinations of the head-class samples/videos in the batch using Mixup, which reduces overfitting on few-shot classes.

\begin{table*} [!t]
        \setlength\extrarowheight{1pt}
        \centering
        \caption{Summary of data balancing approaches to long-tailed learning.}
        \label{tab:databalsummary}
        \small
        \setlength\tabcolsep{2pt}
        \resizebox{\textwidth}{!}{
        \begin{tabular}{m{1.3cm}<{\centering}||m{3.8cm}<{\centering}|m{2.3cm}<{\centering}|m{4.5cm}|m{11.5cm}}
            \hline
             \centering \multirow{2}{*}{\begin{tabular}[c]{@{}l@{}}\quad Sub-\\Category\end{tabular}} &
             \multicolumn{2}{c|}{\centering Representative Methods} &
             \centering \multirow{2}{*}{Rebalancing Strategy} &
             \begin{minipage}{6cm} \centering \multirow{2}{*}{Common Features of the Methods} \end{minipage} \TBstrut\\\cline{2-3}
             & \centering Ref & \centering Venue \& Year & & \TBstrut\\
             \hline
                \multirow{4}{*}{\rotatebox[origin=rc]{90}{Data Resampling}} & 
                                    \centering \multirow{4}{*}{\cite{Shi0XL23}} & 
                                    \centering \multirow{4}{*}{\begin{tabular}[c]{@{}l@{}}\quad NeurIPS 2023\end{tabular}} &
                                    \multirow{4}{*}{\begin{tabular}[c]{@{}l@{}}Class-wise sampling or \\Uniform sampling\end{tabular}} &
                                    \multirow{4}{*}{\begin{tabular}[c]{@{}l@{}}Artificially balance data during model training:\\ 1) undersampling sample with low frequency on the majority (head).\\ 2) oversampling sample with high frequency on the minority (tail), or sample\\ repeatedly on the minority class samples.\end{tabular}}\TBstrut\\
                                    & & & &\TBstrut\\
                                    & & & &\TBstrut\\
                                    & & & &\TBstrut\\
                                    \hline
                                    \hline

                \multirow{8}{*}{\rotatebox[origin=rc]{90}{Data Augmentation}} &
                                    \centering Rand Augment\cite{RandAugment} & 
                                    \centering NeurIPS 2020 &
                                    \multirow{8}{*}{\begin{tabular}[c]{@{}l@{}}1) Image transformation\\2) Employ memory bank to\\ store minority class features\\3) Borrow from external\\ object-centric datasets\end{tabular}} & 
                                    \multirow{8}{*}{\begin{tabular}[c]{@{}l@{}} 1) Object recognition perform data augmentation\cite{RandAugment} based on image\\ transformation, or interpolation\cite{mixup,cutmix}.\\ 2) Object detection use memory bank to store RoI feature and boxes over\\ the past batches, or borrow and stitch external object-centric Image sets to\quad\quad\quad\\ create Scene Images, such as MosaiCOS\cite{MosaicOSLT}.\end{tabular}} \TBstrut\\\cline{2-3} 
                                                                & 
                                    \centering Mixup\cite{mixup} &
                                    \centering ICLR 2018 & &\TBstrut\\\cline{2-3}
                                                                & 
                                    \centering UniMix\cite{UniMix} &
                                    \centering NeurIPS 2021 & &\TBstrut\\\cline{2-3}
                                                                & 
                                    \centering CutMix\cite{cutmix} &
                                    \centering CVPR 2019 & &\TBstrut\\\cline{2-3}
                                                                & 
                                    \centering MetaSAug\cite{MetaSAug} &
                                    \centering CVPR 2021 & &\TBstrut\\\cline{2-3}
                                                                & 
                                    \centering MosaiCOS \cite{MosaicOSLT} &
                                    \centering ICCV 2021 & &\TBstrut\\\cline{2-3}
                                      & 
                                    \centering CUDA \cite{CUDA} &
                                    \centering ICLR 2023 & &\TBstrut\\\cline{2-3}
                                      & 
                                    \centering DODA \cite{DODA} &
                                    \centering ICLR 2024 & &\TBstrut\\
                                    \hline
                                    \hline
                \multirow{6}{*}{\rotatebox[origin=rc]{90}{Data Synthesis}} &
                                    \centering RSG\cite{RSG} & 
                                    \centering CVPR 2021 &
                                    \multirow{6}{*}{\begin{tabular}[c]{@{}l@{}}1) Cluster centroid estimation\\2) Class-specific distribution\\ variance estimation\\3) Feature-level synthesis for\\ minority samples\\4) Using foundation models \end{tabular}} &
                                    \multirow{6}{*}{\begin{tabular}[c]{@{}l@{}}Estimate the variance/covariance or the geometry of the data distribution of\\ different classes in the feature space, then:\\ 1) enforce the variance/covariance of a tail class to be the same as the\\ averaged variance/covariance of the head classes.\\ 2) transfer the geometry of the head classes to the tail classes. \\ or use foundation models for image generation. \end{tabular}}\TBstrut\\\cline{2-3} 
                                                                & 
                                    \centering M2m \cite{m2m} &
                                    \centering CVPR 2020 & &\TBstrut\\\cline{2-3}
                                                                & 
                                    \centering LEAP\cite{LEAP} &
                                    \centering CVPR 2020 & &\TBstrut\\\cline{2-3}
                                                                & 
                                    \centering GistNet\cite{GistNet} &
                                    \centering ICCV 2021 & &\TBstrut\\\cline{2-3}
                                                                & 
                                    \centering DisRobuLT\cite{DisRobuLT} &
                                    \centering ICCV 2021 & &\TBstrut\\\cline{2-3}
                                     & 
                                    \centering LTGC \cite{LTGC} &
                                    \centering CVPR 2024 & &\TBstrut\\
                                    \hline
                                    \hline
        \end{tabular}}
    \end{table*}

\textbf{Cutmix and its variants}. Cutmix \cite{cutmix} is another image data augmentation strategy that replaces a randomly selected region in an image with a patch from another image during image mixing. The soft label of the generated image can be obtained in the same way as Mixup, but the weights to be used in the linear interpolations should be in proportion to the areas of the patches in the generated images. Based on the idea of Cutmix, the Context-Rich Minority Oversampling (CMO) method \cite{CMO} creates synthetic minority images by simply pasting the real minority samples (as the foreground) onto the majority samples (as the background). This way, the rich contexts can be transferred from the majority samples to the synthetic minority samples, making them more diversified for training LTL models. \cite{Shi0XL23} proposes a new augmentation method that first extracts the rich background images from the head classes using activation maps then pastes the tail class images upon them to generate tail-class images with diverse contexts. Instead of arbitrarily pairing the foreground and background images like \cite{CMO}, OTmix \cite{OTmix} considers the semantic distances between the foreground and background images/distributions and selects the most suitable background image to generate more meaningful samples.

\textbf{Other augmentation methods}. ISDA \cite{ISDA} is a data augmentation method that produces diversified augmented samples by translating features along semantically meaningful directions (e.g. colour, backgrounds, visual angles). MetaSAug \cite{MetaSAug} extends ISDA to the long-tailed setting by learning the semantic directions via meta-learning. Major-to-minor Translation (M2m) \cite{m2m} iteratively modifies (translates) a majority sample until it can be confidently classified as a minority sample by a pre-trained classifier built upon the imbalanced dataset.  M2m  leverages the richer information of the majority samples and does not overly use the minority samples, which can thus learn more generalisable features for the minority classes. \cite{DODA} reveals that existing data augmentation methods have side effects since they sacrifice the performance of certain classes. They propose DODA that dynamically maintains class-wise data augmentation preference list by comparing the number of correctly predicted samples of each class in the current and previous epochs. 

\textbf{Augmentation methods for long-tailed object detection}. In \cite{oksuz2020imbalance}, the authors present a comprehensive review on the class imbalance problems in object detection. They summarize four types of imbalance problems in object detection: class imbalance, scale imbalance, spatial imbalance, and objective imbalance. \cite{pRoI} analyzes the issue of Intersection over Union (IoU) distribution imbalance in two-stage object detection and proposes the positive Region of Interest (pRoI) generator, which increases the number of positive examples, especially for those with greater IoU values, and improves the final object detection performance.

The authors in \cite{RIO} point out that, for long-tailed object detection, image-level resampling alone is not enough to yield a sufficiently balanced distribution at the object level. They thus propose a joint resampling strategy, termed \textit{RIO} (Resampling at Image-level and Object-level), which augments the minority classes in the current batch with the object-centric RoI features and box coordinates stored in memory over the past batches. MosaicOS \cite{MosaicOSLT} is another augmentation method that generates pseudo scene-centric images by stitching object-centric images (i.e. ImageNet) for long-tailed object detection. The difference between RIO and MosaicOS is that RIO uses a memory bank to reuse the object-centric RoI features and box coordinates over the past batches; in contrast, MosaicOS straightforwardly generates synthetic scene images by stitching several object-centric images into one image.

\subsubsection{Data Synthesis}
While data augmentation techniques generate synthetic images based on image transformations, data synthesis approaches utilize more complex techniques for image generation, including GANs \cite{qastaaai2023}, distribution estimation/transfer  \cite{RSG,DisRobuLT,LEAP,GistNet}, pre-trained vision/vision-language models (foundation models \cite{FL1,FL2,FL3,FV1}), etc. 

\textbf{GAN based data generation}. \cite{UTLO} finds that information at the lower resolutions tends to be class-independent thus can be shared by all the classes, while class-specific features are usually unveiled at higher solutions. Inspired by such observations, in GAN-based image synthesis for long-tailed data, it proposes to calculate the unconditional (class-independent) objective for low-resolution images, while injecting conditional information at higher-resolution images.

\textbf{Distribution based data synthesis}. The authors in \cite{DisRobuLT} point out that estimating the centroids of the tail classes is much noisier because they have few samples and propose \textit{DisRobuLT}, which optimizes against the worst-case centroids within a safety hyper ball around the empirical centroid. Rare-class sample generator (RSG) \cite{RSG} is a data synthesis method which obtains the feature displacement of a real frequent-class sample by subtracting its feature vector with the closest cluster centre of the same class to remove the class-relevant information and obtain the divergence value. It next adds the above divergence value of the frequent-class samples directly to the feature vector of each rare-class sample to create synthetic minority samples. The authors in \cite{LEAP} propose LEAP, which first uses the angles of the cosine similarity between each feature vector and the centre of the corresponding class centre to model a Gaussian distribution for each class, next forces the variance of the Gaussian distribution for each tail class to be the same as the averaged variance of the head classes, then randomly selects feature vectors using the new distribution of each tail class. LEAP enlarges the variance of the feature distribution of each tail class so that they become more separated from other classes. GistNet \cite{GistNet} also proposes to share the geometry of the head classes with the tail classes. However, unlike LEAP, it uses an alternative implementation of the Softmax function by adding the weight vector with a shape parameter that contains the geometry information shared by all the classes. Since the head classes tend to dominate the optimization of the shared parameters, the geometry of the head classes will be naturally transferred to the tail classes.  Overall, RSG forces the intra-class variation information of the tail classes to be the same as the head classes to generate synthetic samples for the tail classes, while DisRobuLT estimates the centroids of the tail classes using the distributional robustness theory. LEAP uses the averaged variance value of all the head classes as the variance of the tail classes to model the distribution of each tail class, whereas GistNet proposes to learn the shared geometry parameters from the head classes and then transfer them to the tail classes.

\textbf{Foundation model based data generation}. With the rise and prosperity of large pre-trained models/foundation models, e.g.,  CLIP \cite{CLIP}, GPT-4V and DALL-E \cite{DAll-E}, researchers  attempt using such large pre-trained models to generate synthetic data to address the long-tail problem. In \cite{LTGC}, the authors leverage Text-to-Image (T2I) model \cite{T2ISurvey} to generate diverse images for the tail classes, then refine them by  measuring the cosine similarity between the  image and the corresponding text features extracted via CLIP \cite{CLIP}.


\subsection{Neural Architecture}
Neural architecture design, i.e., devising an effective neural network structure/model  for a specified task, long with loss function design, are the two preliminary approaches to improving the overall performance of a neural network (model). Along this line, related methods can be divided into subcategories of decoupled learning framework, early-exit framework, knowledge distillation framework, dual streams inference framework, and multi-expert (ensemble) learning framework.

\begin{figure}[h!]
\centering
\resizebox{1.0\columnwidth}{!}
{
\includegraphics[page=1,width=0.9\columnwidth]{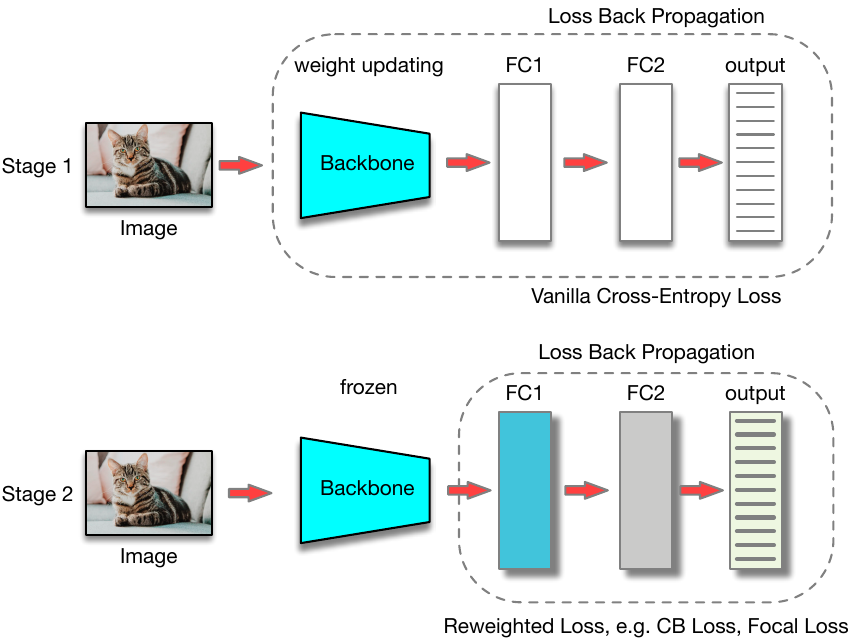}
}
\caption{Neural architecture design for improving the performance of long-tailed learning.}
\label{fig:neuralarch}
\end{figure}

\subsubsection{The Decoupled Learning Framework}
\textit{Decouple} \cite{decouple} is a seminal work that reveals representation learning and classifier training should be carried out separately (i.e. decoupled) for improving long-tailed recognition. Specifically, as shown in Figure \ref{fig:neuralarch}, the proposed decoupled LTL framework first adopts random sampling (instance-based sampling) and the Cross-Entropy loss for training a conventional feature extraction module in the first stage. It then freezes the backbone and uses class-balanced (class-wise) sampling to retrain the classifier in the second stage. Under the same decoupling framework as  \cite{decouple}, \cite{CBD21} proposes to train several teacher models upon batches built via instance sampling in the first stage so that features extracted by different models can be more complementary, then distils the knowledge from the teachers to a student model. \cite{MarrakchiMB21} also adopts the decoupling framework to address the class imbalance issue in medical image datasets, the difference is that it utilizes contrastive learning \cite{SimCLR} to learn the feature space in the first stage. In \cite{balornot}, the authors use class-balanced sampling and conventional random sampling simultaneously to construct mini-batches of data for the feature extraction step.  They reveal that random sampling avoids the risk of overfitting on the tail classes, while class-balanced sampling prevents the training process from being dominated by head classes. 

The above methods have a clear boundary between the two decoupled learning stages. In Bilateral-Branch Network (BBN) \cite{BBN}, the authors propose to use an annealing factor to gradually transits between the two stages, which can adjust the model to learn the global features from the original distributions first, then gradually transit to the tail data modeling and discrimination learning. In Domain Balancing (DB) \cite{domainbalance}, the authors devise a soft gate to switch between a conventional feature learning module and a tail-class favoured feature learning module. For each tail-class image, the features from the two modules are integrated to improve the LTL performance. Hybrid-SC \cite{wangcontrastive} proposes a hybrid network structure composed of a supervised contrastive learning branch to learn better feature representations from the long-tailed data and a  conventional CNN branch with a cross-entropy loss to learn the classifiers. Like BBN, it smoothly transits between the two branches via a  cumulative learning adaptor. The main difference between Hybrid-SC and BBN is that the former uses supervised contrastive learning in the first stage, which can learn better features. In \cite{DT2}, the authors reveal that the inference of visual relationships in a scene faces the problem of the dual long-tailed distributions of the entity (object) and predicate (relation) classes. To address this problem, they propose the DT2 model, which is based on an iterative decoupled training scheme. Like BBN, DT2 first trains (warms up) the model on the original data with uniform sampling, then fixes the backbone part and fine-tunes the classifier on the (remaining) network with the proposed Alternating Class Balanced Sampling (ACBS) method to capture the interplay between the long-tailed distributions of entities and predicates using knowledge distillation.  

\cite{DisAlign} points out that the performance bottleneck of the two-stage (decoupling) learning framework seems to be the classifier retraining issue in the second stage. It thus proposes \textit{DisAlign} for confidence calibration to adjust the decision boundary in the second (classifier retraining) stage. In \cite{clsinceccv2022}, the authors propose the long-tailed class incremental learning problem and present a preliminary method under the two-stage LTL framework that adds a layer of learnable scaling weights to integrate the outputs of the classifier heads for class-incremental learning.

\subsubsection{The Early-Exit Framework} \label{earlyexitfra}
In \cite{earlyexit}, the authors point out that existing approaches to long-tailed recognition reweight samples only by class size, but without considering the fact that there might be easy examples in the minority classes that get incorrectly up-weighted and hard examples in the majority classes that get erroneously down-weighted. They thus employ the early exiting framework (ELF) to early-exit easy examples in the shallow layers of the network to focus on the hard examples in each class in the deeper layers. In implementation, ELF augments a backbone network with several auxiliary classifier branches. During training, it finds and early-exits the correctly and confidently classified examples at the earliest possible exit branch, while harder examples will exit later in the network, sustaining their impact on the overall loss. Similarly, \cite{HAR} proposes the Hardness Aware Reweighting (HAR) framework, which also adopts the early-exiting strategy in the training process. Specifically, HAR augments a backbone network with auxiliary classifier branches, and samples that are confidently and correctly classified at a branch will early exit from that branch.

\subsubsection{The Knowledge Distillation Framework}
In \cite{SSDLT}, the authors propose Self-Supervised Distillation for long-tailed visual recognition (SSD-LT). Adopting the decoupled learning framework, it trains two heads for knowledge distillation, which are the self-distillation head and the classification head, respectively. It thus owns the merits of rebalanced sampling and decoupled training strategy and is reported to obtain state-of-the-art performance in long-tailed recognition. The authors in \cite{DiVE} propose the ``Distill the Virtual Examples'' (DiVE) pipeline for long-tailed recognition. In DiVE, a teacher model is first trained for the long-tailed task with the existing methods, and the predictions from such a teacher model are used as virtual examples. It then uses knowledge distillation to transfer the knowledge from the teacher model to the student model.

\cite{DeiT-LT} discusses the use of ViT/Transformers for long-tailed learning. It proposes DeiT-LT which distills knowledge from a CNN teacher to a ViT backbone/student using out-of-distribution images generated via CutMix and Mixup, in which it reweights the distillation loss to enhance the focus on the tail classes.  Recently, for long-tailed object detection, RichSem \cite{RichSem} adds a new branch to the detection framework for distilling semantics from CLIP, which can extract rich semantics for the bounding box patches within the image. 

\subsubsection{The Dual Stream Inference Framework} \label{dualinfer}
\cite{MLLT} designs a neural network architecture with two branches: one branch trains a baseline model on the original long-tailed distribution, while the other trains a second model upon the artificially balanced distributions. They add a cross-branch consistency loss based on their logit difference to enforce the two branches to train the multi-label visual recognition model collaboratively. For long-tailed face recognition, Center-dispersed (CD) Loss \cite{zhong2019unequal} first splits the original data into head-class data and tail-class data, then feeds it into the ``unequal-training'' framework that consists of a  backbone and two learning branches for the head classes and tail classes, respectively. The first learning branch is based on the head-class data and is used to stabilise face representation, supervised by a noise-resistant loss. The second learning branch is based on the tail-class data and is used to enhance the inter-class discrimination capability. The training of the framework alternates between the two branches. One difference between \cite{MLLT}  and \cite{zhong2019unequal} lies in that the latter trains two models on the split data (head-classes data and tail-classes data) separately, whereas the former trains two models on the original/complete data.

In SimCal \cite{devilltseg}, the authors investigate instance segmentation over data with long-tail distributions. Upon Mask R-CNN, their proposed method makes three main modifications: i)  it changes the inputs of the original dataset with batches of class-balanced samples, then selects class-balanced object proposals for each class in the region proposal generation (RPN) stage;  ii) in the classification/segmentation stage, it trains a calibrated classifier with Mask R-CNN on the above proposals; iii) in the inference step (after the training is over), it employs the calibrated classifier and the classifier trained on the original dataset using the vanilla Mask R-CNN algorithm to predict the tail and head classes, respectively. Such a dual-head inference framework yields improved performance on the tail classes while still maintaining similar performance on the head classes.

\subsubsection{The Multi-Expert Learning Framework} \label{multiexpert}
Ensemble learning methods aim to utilize multiple experts/models to make more accurate predictions. Representative ensemble learning methods for LTL include ACE \cite{ACE}, RIDE \cite{RIDE},  CBD \cite{CBD21}, LFME \cite{LFME}, BalPoE \cite{BalPoE}, SHIKE \cite{SHIKE}, MDCS \cite{MDCS} and LGLA \cite{LGLA}. 

The authors in \cite{ACE} propose ACE, a multi-expert ensemble network for long-tailed recognition. It aims to achieve biased representation learning and unbiased classifier training in a single stage. The class indices are first ordered descendingly by their number of samples, the first branch (expert) trains a classifier/model with the samples of all the classes; the next branch (expert) first eliminates the current head class, then trains a new classifier with the samples of the rest classes, and so on. Therefore, the number of classes handled by different classifiers decreases monotonically, and the medium-shot or tail classes will have chances to dominate a classifier, thus mitigating the bias towards the majority classes. Overall, ACE can simultaneously improve the recognition accuracy of both the head and tail classes. RIDE \cite{RIDE} trains multiple recognition models/experts on randomly sampled subsets and uses their logits mean to make predictions. It also devises a routing algorithm to skip the inferences of certain experts. CBD \cite{CBD21} also trains multiple recognition models/experts on randomly sampled subsets, it proposes to distil the knowledge from the multiple teachers/experts into a single student model. Similarly, LFME \cite{LFME} also propose to distil the knowledge from multiple teachers into a unified/single student model. LFME differs from CBD in that the teacher models of the LFME are trained over non-overlapped partitions of classes and samples, while CBD trains multiple teacher models over randomly sampled subsets which may have overlapped classes/samples. Compared to CBD and LFME, ACE \cite{ACE} trains cascaded models/classifiers by iteratively removing the dominating group of classes and the corresponding samples before training the next teacher model, while RIDE  straightforwardly utilize logits mean/average instead of knowledge distillation to ensemble/ally the teacher models. 

The authors in \cite{BalPoE} propose BalPoE, which is also an ensemble learning framework composed of logit-adjusted experts \cite{logitlongtail}. They prove that the ensemble is Fisher-consistent for minimizing the balanced error. SHIKE \cite{SHIKE} is a knowledge distillation-based multi-expert ensemble learning framework for long-tailed recognition. The authors find that shallow features of a network (or a shallow network) can better represent the tail classes. They thus propose aggregating the deep features in different experts/branches with different layers of shallow features, respectively. Next, they use knowledge distillation techniques to let the experts learn from each other. Like other ensemble learning algorithms, the averaged predictions from the multi-experts will then be used as the final prediction. Nested Collaborative Learning (NCL) \cite{NCL} also trains multiple experts on all the categories and some of the hard categories, respectively. It then uses knowledge distillation to learn the multiple experts collaboratively. MDCS \cite{MDCS} is a multi-expert learning framework that introduces a diversity loss (which is essentially the Balanced Softmax loss \cite{BALMS}) to control experts' focus on different categories with an adjustable weight to improve the diversity of experts, and performs  self-distillation between weakly and strongly augmented logits sets for each expert to distill the richer knowledge of weakly augmented instances to reduce model variance. 

Considering the fact that the class distributions of test data are not necessarily uniform but might be long-tailed or even inversely long-tailed, the authors in  \cite{SADE} propose SADE which is a test-agnostic long-tailed recognition approach that trains three experts/models on uniform, long-tailed, and inversely long-tailed class distributions separately, then weighted aggregates them at the inference stage. DirMixE \cite{DirMixE} is also an multi-expert learning approach, which uses the Monte Carlo method to estimate the mean and semi-variance of a meta-distribution for capturing both the global and local variations to adapt to test-agnostic distributions. LGLA \cite{LGLA} trains multiple experts on the entire dataset to extract more discriminative features for different subsets, which uses logit adjustment strategy to enlarge the discrepancy among the subsets. It also learns an expert excelling at inversely long-tailed distribution, next ensembles these experts to achieve superior prediction performance.

\cite{LPT} proposes to leverage the powerful discrimination ability of the large-scale pre-trained vision models \cite{CLIP,FV1} to tackle long-tailed classification directly. It proposes LPT which first adapts the pre-trained model to the target domain using shared prompt tuning, next adopts the divide-and-conquer strategy for long-tailed recognition by using the pre-trained model and shared prompt to generate a  group-specific prompt set for each group of classes. 

Besides, lately, the Transformer architecture  \cite{TransformerSurvey} is being increasingly utilized in LTL \cite{LiVT}. Transformer based LTL methods, such as  LTGC \cite{LTGC} and VL-LTR \cite{VL-LTR} reported significantly better performance than CNN based methods in long-tailed recognition, while Step-wise \cite{Dong_2023_ICCV}  which adopts the CNN-Transformer combined architecture, also showed superior performance in long-tailed object detection. 

\subsection{Feature Enrichment}

\begin{figure}[h!]
\centering
\includegraphics[page=1,width=1.0\columnwidth]{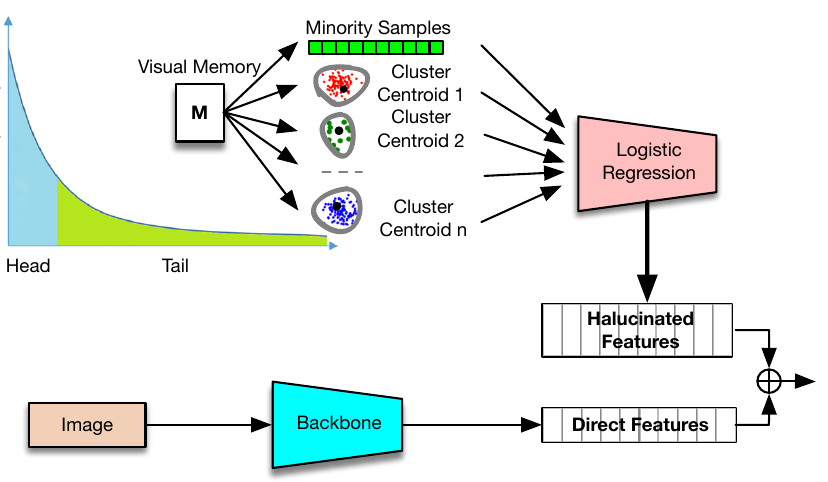}
\caption{Feature enrichment approaches to improving long-tailed learning performance.}
\label{fig:featureAug}
\end{figure}

Feature enrichment methods enrich/augments features of the tail classes  using fusion-based approaches \cite{openlongtail,RAC,EquiLTOD,domainbalance,ChuBLL20,FeaHalluOneShot},  perturbation based approaches \cite{Procrustean,FASA}, large pre-trained models/foundation models\cite{CLIP,ALIGN,VL-LTR}, etc. 

\subsubsection{Fusion Based Feature Enrichment}

As depicted in Figure \ref{fig:featureAug}, the Retrieval Augmented Classification (RAC) method \cite{RAC} uses an external memory module to store features of the tail samples explicitly. Next, for each input image, it finds in the memory bank the Top-K most similar features of the same class, which are then fed into a BERT-like encoder to be transformed into logits, then fused with the logits derived from a conventional backbone. Likewise, the authors in \cite{PFENet} propose PFENet, a prior guided feature enrichment network that enriches query features with support features and prior masks to overcome the spatial inconsistency between the query and support samples in few-shot segmentation. LOCE  \cite{EquiLTOD} also utilizes a memory module to store features of the tail-class samples for augmenting the features of the tail classes, which works collaboratively with the proposed  Equilibrium Loss (EBL) that adjusts the margins for the tail classes in long-tailed detection. In OLTR \cite{openlongtail}, the authors combine the features extracted by a conventional CNN pipeline and the features stored in an additional memory module to generate features of the tail classes, which can transfer knowledge between head and tail classes, and increase the sensitivity of the model to novel classes.

\cite{FeaHalluOneShot} proposes to disentangle image features into  class-agnostic features (a.k.a. category-related/class-relevant) and class-generic features (a.k.a. category-independent/class-irrelevant)  for  one-shot image generation. It uses an auxiliary memory module to store class-generic features. The class-agnostic features from the input image and the class-generic features from the memory module are then fused and sent to the generator of a vanilla GAN for image generation. Likewise, in \cite{ChuBLL20}, the authors point out that some features are shared by the head and tail classes, while other features are class-agnostic (unique to certain classes). The authors thus propose to transfer the class-generic features from the head to the tail classes. \cite{H2T} propose to augment tail classes by grafting the diverse semantic information from head classes, where it replaces a portion of feature maps from tail classes with those belonging to head classes. Such fused features substantially enhance the diversity of tail-class representations. To prevent the neural networks from fitting the head classes first, MFW \cite{Procrustean} weakens features of the head classes by mixing them with features of other classes in the same mini-batches but without altering the corresponding labels to equalize the training progress across classes.

CoSSL \cite{CoSSL} devises a tail-class feature enhancement mechanism that gives a higher probability to the tail classes when blending features of unlabeled and labelled data.

\subsubsection{Perturbation Based Feature Augmentation}
The authors in \cite{FASA} propose the \textit{FASA} method for long-tailed instance segmentation that generates class-wise virtual features by random perturbation over a Gaussian distribution on the real samples of each class. 

CC-SAM \cite{CC-SAM} proposes to improve the generalization capability of the LTL models by robust training against class-conditioned perturbations on the features of the last convolutional layer. Its idea is similar to that of LogitAdjust \cite{logitlongtail}. The main difference is that LogitAdjust performs perturbations over logit values in the output layer, whereas CC-SAM adds noises to the features in the last convolutional layer (the last hidden layer).

\subsubsection{Foundation Model Based Feature Enrichment}

Recently, foundation models such as CLIP \cite{CLIP} and ALIGN \cite{ALIGN} have demonstrated impressive representation and generalization capabilities. In \cite{VL-LTR}, researchers have  explored using such foundation models for feature-level enhancements. They simultaneously utilize the visual and linguistic representations extracted by the foundation models CLIP \cite{CLIP} for long-tailed recognition. For long-tailed object detection, RichSem \cite{RichSem} adds a new branch to the detection framework for distilling knowledge from  CLIP, which can extract rich visual and linguistic semantics to enhance the representations of the objects within the image.

\subsection{Logits Adjustment} \label{subsec:logitadjust}
In the output layer, logits contain the raw prediction values of the neural network, which will be used as the input to the Softmax activation function to turn the logit values into a probability distribution over the input classes. Methods that fall into the logits adjustment category either optimize the inherent process in deriving the logit values  \cite{IR-Softmax,kim2020adjusting,BMSE}, or calibrate the logit values in a post hoc manner \cite{GCL,DebiasPL}. It also includes softmax margin learning methods \cite{marginimloss,equaloss,AutoBalance,BALMS} that enforce margins into the logit values so as to influence the resulting Softmax classifier to have larger classification margins \cite{marginlearningsurvey}. We note that, sometimes it is challenging to clarify the differences between logits calibration and Softmax margin learning. For instance, LogitAdjust \cite{logitlongtail} can be considered to be belonging to both subcategories. In Table \ref{tab:Loss7}, we summarize the formulas of representative methods in this category.

\begin{table*}
        \centering
        \caption{Summary of  logits adjustment methods and loss reweighting methods for long-tailed learning.}
        \label{tab:Loss7}
        \small
        \setlength\tabcolsep{2pt}
        \resizebox{\textwidth}{!}{
        \begin{tabular}{|m{0.5cm}<{\centering}||m{2.2cm}<{\centering}|m{1.3cm}<{\centering}|m{2cm}<{\centering}|m{7cm}<{\centering}|m{8cm}<{\centering}|}
            \hline
             &\centering Method \& Ref. & \centering Venue & \centering Task & \centering Loss Function & \begin{minipage}{8cm} \centering Notes/Remarks \end{minipage}\\
                \hline
                                \multirow{8}{*}{\rotatebox[origin=rc]{90}{Logits Adjustment  \quad \quad \quad \quad \quad}} 
                                    & 
                                    \centering Balanced MSE\cite{BMSE} & CVPR 2022 &
                                    \centering Visual regression & \centering$-\log_{}{(\frac{\exp(-\left \| p_{i} - y_{i} \right \|_{2}^{2}/\tau )}{\sum_{j=1, j\neq i}^{K}\exp(-\left \| p_{j} - y_{j} \right \|_{2}^{2}/\tau )})}$ &
                                    $y_{i}$ is the ground-truth value of class i in the one-hot vector, $p_{i}$ is the predicted probability on class i, $\tau$ is the temperature hyper-parameter. \\\cline{2-6}
                                    & 
                                    \centering LDAM Loss\cite{marginimloss} & NeurIPS 2019 &
                                    \centering Object classification & \centering$-\log_{}{(\frac{n_{i}^{\frac{1}{4}}\cdot\exp(z_{i})}{\sum_{j=1}^{K} n_{j}^{\frac{1}{4}}\cdot\exp(z_{j})})}$ &
                                    It employs a data resampling method that estimates the optimal sampling rates for different classes, $n_{i}$ is the number of samples in the class of the input sample.\\\cline{2-6}
                                    &                       
                                    \centering EQL Loss\cite{equaloss} & CVPR 2020 &
                                    \centering Object detection and recognition & \centering$-\log_{}{(\frac{\exp(z_{i})}{\sum_{j=1}^{K}w_{j}\cdot\exp(z_{j})})}$ &
                                    EQL Loss is short for Equalization Loss, $w_{j}$ is a factor that aims to ignore or down-weight  the gradients of other tail classes, while still keeping the gradients for the samples of the head classes.\\\cline{2-6}
                                    & 
                                    \centering AutoBalance\cite{AutoBalance} & NeurIPS 2021 &
                                    \centering Object classification & \centering$-\omega_{i}\cdot\log_{}{(\frac{\exp(s_{i}\cdot z_{i}+m_{i})}{\sum_{j=1}^{K}\exp(s_{j}\cdot z_{j}+m_{j})})}$ &
                                    $s_{i}$, $m_{i}$ are the multiplicative and additive margin parameters, $\omega_{i}$ is the class-wise reweighting factor for the group of classes having similar number of samples.\\\cline{2-6}
                                    & 
                                    \centering LogitAdjust\cite{logitlongtail} & ICLR 2021 & 
                                    \centering Object classification & \centering$-\log_{}{(\frac{\exp(z_{i}-\tau\cdot m_{i})}{\sum_{j=1}^{K}\exp(z_{j}-\tau\cdot m_{j})})}$ &
                                    $m_{i}$ is a margin parameter for class $i$, $\tau$ is a hyper-parameter (temperature).\\\cline{2-6}
                                    & 
                                    \centering BALMS Loss \cite{BALMS} & NeurIPS 2020 &
                                    \centering Object classification & \centering$-\log_{}{(\frac{\exp(z_{i}-m_{i})}{\exp(z_{i}-m_{i})+\sum_{j=1, j\neq i}^{K}\exp(z_{j})})}$ &
                                    The margin $m_{i}=\frac{P}{n_{i}^{\sfrac{1}{4}}}$ is reversely proportional to the quartic root of the number of samples of the class, $P$ is a hyper-parameter.\\\cline{2-6}
                                     & 
                                    \centering C2AM \cite{C2AM} & CVPR 2022 &
                                    \centering Object detection & $-\log_{}{(\frac{\exp(s\cdot z_{i})}{\exp(s\cdot z_{i})+\sum_{j=1, j\neq i}^{K}\exp(s\cdot z_{j}+m_{ij})})}$\centering  &
                                    $z_{i} = \cos \theta _{i}$, $m_{ij} = \left | \frac{\alpha }{\pi }\log_{}{\frac{\left \| w_{i}  \right \|_{2} }{\left \| w_{j}  \right \|_{2}}}   \right |$, $\left \| w_{i}  \right \|_{2} $ is the L2 norm of the corresponding weight vector for class i, $\alpha$ and $s$ are hyper-parameters. \\\cline{2-6} 
                                    & 
                                    \centering GCL\cite{GCL} & CVPR 2022 &
                                    \centering Object recognition & \centering $-\log_{}{(\frac{\exp(s\cdot(z_{i}-m_{i}\cdot\epsilon_{i}))}{\sum_{j=1}^{K}\exp(s\cdot(z_{j}-m_{j}\cdot\epsilon_{j}))})}$&
                                    $z_{i} = \cos \theta _{i}$, which is the cosine of the angle between the logit vector and the corresponding weight vector for the class of the input sample.  $m_{i} = \log_{}{\frac{n_{max}}{n_{i}}}$. $\epsilon_{i} = \mathcal{N}(0,1)$, which is random decimal. $s$ is a hyper-parameter. \\
                                    \hline
                                    \hline
                \multirow{7}{*}{\rotatebox[origin=rc]{90}{\centering Loss Reweighting \quad \quad \quad \quad \quad \quad \quad \quad \quad \quad}} 
                                    &
                                    \centering CB Loss\cite{cbloss} & CVPR 2019 & 
                                    \centering Object classification & \centering$-\frac{1-\beta}{1-\beta^{n_{i}}}\cdot\log_{}{(\frac{\exp(z_{i})}{\sum_{j=1}^{K}\exp(z_{j})})}$\ & $\beta=0.999$ or other values, $n_{i}$ is the number of samples in the class which the input sample belongs to. CB Loss inversely reweights the loss of a sample by the number of samples in that class.\\\cline{2-6}
                                     & 
                                    \centering Range Loss\cite{rangeloss} & ICCV 2017 &
                                    \centering Face classification and recognition & \centering$\alpha\cdot\sum_{i=1}^{}\frac{2}{\frac{1}{d_{1}}+{\frac{1}{d_{2}}}}+\beta\cdot\left|P-d_{3}\right| -\lambda\cdot\log_{}{\frac{\exp(z_{i})}{\sum_{j=1}^{K}\exp(z_{j})}} $ &
                                    $d_{1}$, $d_{2}$ are the two largest intra-cluster distances of the input sample's class, $d_{3}$ is the shortest distance of the centres of any two classes, $P$ is a hyperparameter. \\\cline{2-6}
                                    & 
                                    \centering Seesaw Loss\cite{SeeSaw} & CVPR 2021 &
                                    \centering Instance segmentation & \centering$-\log_{}{(\frac{\exp(z_{i})}{\exp(z_{i})+\sum_{j=1, j\neq i}^{K}s_{j}\cdot\exp(z_{j})})}$ &
                                    Based on the Softmax Cross Entropy loss, but multiplies $s_{j}$ in the denominator, which is a dynamic balancing factor that jointly employs prediction probability and the ratios between the number of samples. \\\cline{2-6}
                                    &
                                    \centering TCR Loss\cite{balancerethinking} & CVPR 2020 & 
                                    \centering Object classification & \centering$-(\omega_{i}+\varepsilon_{i})\cdot\log_{}{(\frac{\exp(z_{i})}{\sum_{j=1}^{K}\exp(z_{j})})}$\ &
                                    TCR Loss is short for the two-component hybrid reweighting method. $\omega_{i}$ and $\varepsilon_{i}$ are obtained via CB Loss \cite{cbloss} and L2RW loss \cite{L2RW}, respectively. \\\cline{2-6}
                                    &
                                    \centering Focal Loss \cite{focalloss} & ICCV 2017 &
                                    \centering Object detection and recognition & \centering$-(1-p_{i})^{r}\cdot\log_{}{p_{i}}$, $ $ $p_{i} = \frac{\exp(z_{i})}{\sum_{j=1}^{K}\exp(z_{j})}$ &
                                    $r$ is the focusing parameter, e.g. $r=2$. Difficult examples will be relatively upweighted $(1-p_{i})^{r}$, while easy examples (those with less prediction loss value) will be relatively down-weighted.\\\cline{2-6}
                                                                & 
                                    \centering IB Loss \cite{IBLoss}& ICCV 2021 & 
                                    \centering Object classification (Instance-wise) & \centering$-\frac{1}{\sum_{j=1}^{K}\left |p_{j}-y_{j} \right| \cdot \sum_{l=1}^{L}\left |h_{l} \right|}\cdot\log_{}{(\frac{\exp(z_{i})}{\sum_{j=1}^{K}\exp(z_{j})})}$&
                                    IB Loss is short for Influence-Balanced Loss, it contains a normal training step and a fine-tuning step. $p_{j}$ is the predicted probability on the class $j$ in the output layer, $h$ is the logit vector. \\
                                    \hline
                                    \hline
        \end{tabular}}
    \end{table*}

\subsubsection{Logits Calibration}

\textbf{Inherent Logits Calibration}. When deriving logits from the last hidden layer to the output layer, IR-Softmax \cite{IR-Softmax} sets the weights for each class as their corresponding class centres in the feature space to avoid the shift between the weights and their centres. The authors in \cite{BMSE} propose \textit{Balanced MSE} that straightforwardly uses the mean square error on each class to replace the corresponding logit element in imbalance regression. 

\textbf{Post hoc Logits Calibration}. In \cite{GCL}, the authors argue that the obscure regions between classes in classification are caused by Softmax saturation. To alleviate this issue, they propose the Gaussian Clouded Logit adjustment (GCL) loss that perturbs the normalized logit with an additive margin that has a random disturbing parameter. In \cite{DebiasPL}, the authors reveal that pseudo-labels in semi-supervised learning are naturally imbalanced. They propose a Debiased Pseudo-Labeling method (DebiasPL) that deducts the probability for each class from the logit. After that, margin learning methods can still be applied. 

Like GCL \cite{GCL}, in BLV \cite{BLV}, category-wise variations are introduced into network predictions (logits) for long-tailed semantic segmentation, where tail classes obtain larger variations by multiplying the random value in a Gaussian distribution with a coefficient that is inversely proportional to the per-class sample size. This simple strategy improves the segmentation performance on the tail categories.

\cite{CDMAD} finds that a classifier trained on long-tailed/imbalanced data produces skewed class probabilities on a solid color image, rather than uniform probabilities, it thus proposes to adjust the classifier's predicted logits on a test sample by subtracting its predicted logits on a white image, leading to improved recognition performance.

Most LTL methods train a model on long-tailed data but evaluate their performance on uniformly distributed data. To address this issue, the authors in \cite{LADE} propose the LAbel distribution DisEntangling (LADE) method, which regularizes the logits so that they can be explicitly disentangled from the source label distribution and adapt to arbitrary target distributions.

\subsubsection{Softmax Margin Learning}
Softmax margin learning algorithms alter the logit values before sending them to the Softmax classifier (Softmax activation function) to improve the discrimination power of the classifiers. Therefore, it can also be considered belonging to the ``Post hoc Logits Calibration'' sub-category. In label-distribution-aware margin loss (LDAM) \cite{marginimloss}, when using the logits to calculate the  Softmax output,  the corresponding logit element for the ground-truth class of the input sample will subtract a class-wise margin parameter, which is reversely proportional to the quadratic root of the number of samples in the class. Equalization Loss (EQL) \cite{equaloss} removes the logit values of other tail classes in the denominator of the Softmax activation function. AutoBalance \cite{AutoBalance} adopts a generalized cross entropy loss with multiplicative and additive margin parameters, and uses a bi-level optimization procedure to learn these parameters on a validation set directly. The overall ideas of BALMS \cite{BALMS} and  LogitAdjust \cite{logitlongtail} are very similar to AutoBalance: the main difference lies in the multiplicative and additive margin parameters used in the Softmax function, as can be seen from Table \ref{tab:Loss7}. Similarly, in \cite{C2AM}, the authors propose the Category-Aware Angular Margin (C2AM) loss for long-tailed object detection, which adds a margin for every non-ground-truth class,  based on the ratio between the weight norms for the ground-truth class and every other class. In \cite{ECM}, the authors deduce that margin-based binary classification provides a closed-form solution for the ideal margin of each object category in long-tailed object detection and propose the Effective Class-Margin (ECM) loss for optimizing margin-based classification.

In \cite{PML}, the authors propose a progressive margin loss (PML) for facial age classification. They propose a framework that consists of an ordinal margin learning head and a variational margin learning head specially tailored to long-tailed data. The weighted sum of the two margins is adopted to make adjusted predictions.

As a short summary, logits adjustment either intervenes the derivation process of logits, which occurs between  the last hidden layer and the output layer, or modifies the existing logit values, primarily by adding different margin parameters. Ultimately, the calibrated logits will be fed into the Softmax Activation Function, resulting in a better-performing Softmax classifier.

\subsection{Loss Function}
\begin{figure}[h!]
\centering
\includegraphics[page=1,width=1.0\columnwidth]{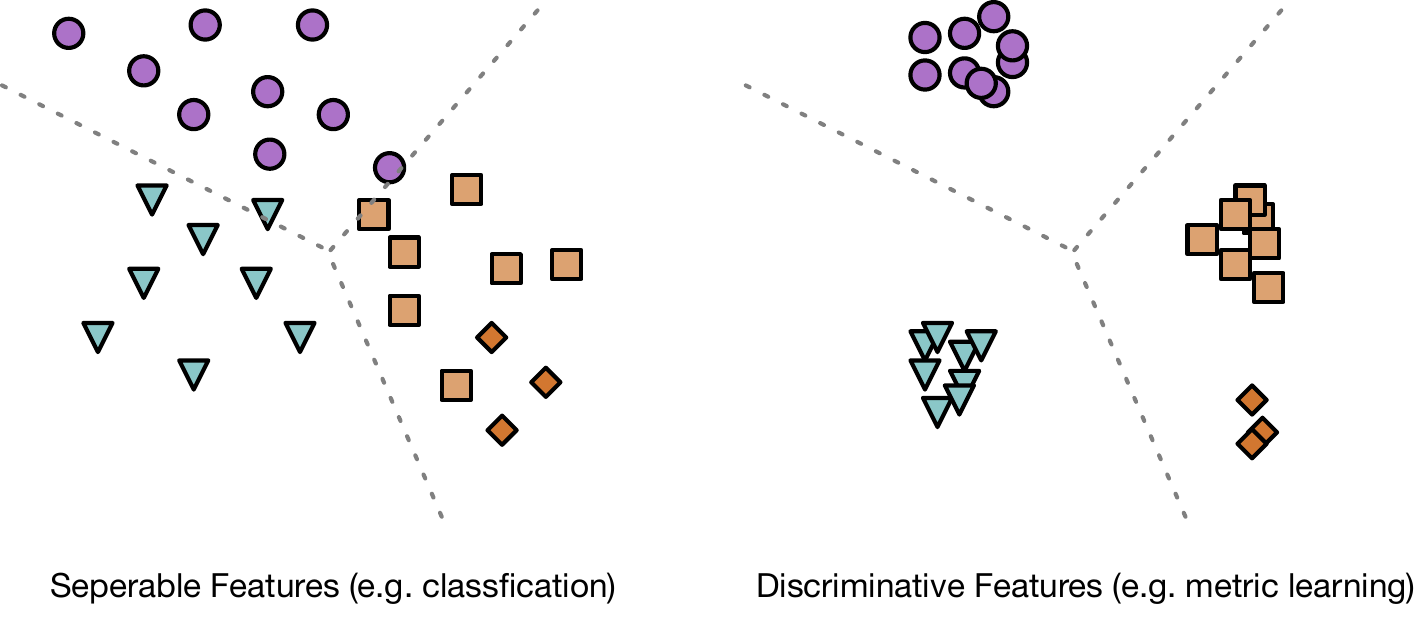}
\caption{Embedding learning approaches to long-tailed learning.}
\label{fig:embedlearn}
\end{figure}

An embedding space, also known as the latent space or latent feature space, is the new space produced by a learning algorithm where semantically similar objects are positioned close together, and dissimilar objects are placed far apart \cite{sanakoyeu2019divide}, as illustrated in Figure \ref{fig:embedlearn}. Roughly speaking, embedding (metric) learning refers to any technique or process that converts an input image into an effective feature representation. Loss function design, along with neural architectures,  are the main approaches to embedding learning and optimization. In fact, the aforementioned feature enrichment techniques, logit adjustment techniques, neural architecture,  as well as training strategies (bells and whistles), are all integral parts of the overall embedding learning process. 

\subsubsection{Loss Function For Embedding Learning} \label{subsec:lossembed}

In this subsection, we will  introduce various loss functions for deep embedding learning in long-tailed settings.

\textbf{Long-tailed  Recognition}. The authors in \cite{hayat2019gaussian}  propose Affinity Loss which jointly performs max-margin classification and feature space clustering for enhancing inter-class separability and reducing intra-class variations. DASO \cite{DASO} generates unbiased pseudo-labels by blending two complementary pseudo-labels obtained via a linear classifier and the similarity computation between the feature vector of a sample and each class centre, respectively. In DLSA \cite{DLSA}, the authors propose to tackle long-tailed recognition by adjusting the label space of the samples via deep clustering. They first cluster the samples according to their likelihood in the feature space. The well-clustered samples are sent to a dedicated Cluster-aided Classifier, which exploits the cluster prior information in classification. The remaining samples are outliers and are sent to the next adjustment module to decrease the training difficulty further. SBCL \cite{SBCL} devises a triplet loss like contrastive loss  for long-tailed recognition. It initially divides the head classes into multiple sub-classes  and enforces a sample to be closer to samples of the same sub-class than other samples in the same class, while being closer to samples of a different sub-class under the same class than samples from different classes in the loss function. GLMC \cite{GLMC} is based upon the Siamese network architecture: for each pair of samples, it respectively uses the CMO \cite{CMO} and Mixup \cite{mixup} methods to create two synthetic images, then uses both the cosine distances between the features of the two images and the classification loss on the two images to optimize the network.  

In \cite{vMFECCV2022}, the authors define the distribution overlap coefficient to quantify the dominance severity of the head classes in the representation space and formulate the inter-class discrepancy and class-feature consistency losses to close the gap. The authors in \cite{SuperDisco} find that the distribution of the super-classes in imbalanced data is more balanced than the original data. They thus propose \textit{SuperDisco} to discover the super-class in a hierarchy of semantic abstraction, then refine the original features of a sample using the corresponding super-class features. 

Contrastive learning  \cite{SimCLR} has recently gained popularity since it can substantially improve the capability of the neural network in representation learning. The authors in \cite{kcl} propose the K-positive Contrastive Loss (KCL) for LTL, which draws k instances from the same class to form the positive sample set for contrastive learning. Based upon KCL, targeted supervised contrastive learning (TSC) \cite{TSC} adds a new loss term between the samples and the predefined optimal class centres to move the samples closer to the centres of their class and away from centres of other classes.  The PaCo \cite{PaCo} loss introduces learnable class-wise centers into the supervised contrastive learning process to make the probability that two samples are a true positive pair become more balanced across different classes, which is beneficial for long-tailed recognition. Model-Aware K-center (MAK) \cite{MAK} sorts the external samples by their empirical contrastive loss expectation value, then selects the Top-N samples to perform K-means clustering and chooses the final samples. The authors in \cite{BCL} propose  Balanced Contrastive Learning loss (BCL) that includes cluster compactness in the contrastive learning process. 

In \cite{wangcontrastive}, the authors design a hybrid network comprising a prototypical supervised contrastive (PSC) learning module and a classifier training module. A curriculum is designed to control the weightings of these two modules during network training.  Motivated by the observation in \cite{hooker2019compressed} that model pruning negatively influences the prediction accuracy on samples of the infrequent classes, the authors in \cite{SDCLR} design the  Self-Damaging Contrastive Learning (SDCLR) framework, which adds a pruned contrastive learning module to the (original/non-pruned) unsupervised contrastive learning module, to amplify the prediction differences between the pruned and non-pruned models and increase the weights of the infrequent class samples and hard samples in the overall loss. SDCLR increases both the overall accuracy and the balanceness of the learned representations.

\cite{CReST} reveals that existing semi-supervised learning methods obtain low recall performance in the minority classes, but the corresponding precision performance is high. It thus proposes \textit{CReST} for  semi-supervised LTL that iteratively samples pseudo-labelled data from the unlabeled set to expand the labelled set, then retrains the semi-supervised learning model. SimPro \cite{SimPro} is a simple probabilistic framework for long-tailed semi-supervised learning, which explicitly separates the estimation process into conditional and marginal distributions using the Expectation-Maximization algorithm.  ABC \cite{abc} attaches an auxiliary classifier head to the representation layer of a semi-supervised learning framework to avoid bias towards the majority classes in LTL. ACR \cite{ACR} uses a dual-branch network to adaptively handle various class distributions of unlabeled data, in which it derives the scaling parameter using the bidirectional KL divergence between the estimated class distribution and three anchor distributions. 

\textbf{Long-tailed  Regression}. When quantizing the imbalanced regression target space into interval classes for imbalance regression, HCA \cite{HCA} proposes a coarse-to-fine hierarchical classification strategy which first uses coarse classifiers to provide the range, next selects the corresponding finer classifier in the range, and achieves improved regression performance.

\textbf{Long-tailed  Object Detection and Segmentation}. The authors in \cite{li2020overcoming} propose the Balanced Group Softmax method (BAGS), which groups the classes by the number of instances, then trains a separate Softmax classifier for each group to reduce the suppression by the head classes. However, finding a proper group partition number for  BAGS could be laborious and tricky. Thus, the authors in \cite{ACSL} devise the Adaptive Class Suppression Loss (ACSL), which utilizes the output confidence to prevent the classifiers of tail categories from being over-suppressed by the samples from head categories, and only the samples from visually confusing categories are kept for discriminative learning. Learning to Segment the Tail (LST) \cite{segtail} also adopts the ``divide-and-conquer'' strategy for long-tailed segmentation. It divides the original data into several balanced parts and devises a knowledge distillation loss which measures the logit difference of the same class in neighbouring parts to preserve knowledge learned in previous parts. \cite{AHR} adds an Adaptive Hierarchical Representation (AHR) loss to the traditional object detection loss for more discriminative embedding learning. It calculates both the coarse-grained classification loss for distinguishing each cluster clearly, and the fine-grained classification loss within each cluster by adopting adaptive margins. In \cite{HarmonicLoss}, the authors point out that, in contemporary deep learning-based object detectors, classification and regression are often optimized independently, which leads to inconsistent predictions with high classification scores but low localization accuracy, and vice versa. They thus propose a harmonic mutual-supervision method which devises a regression-aware classification loss to optimize the classification branch and a classification-aware regression loss to supervise the optimization of the regression branch. \cite{HSSLLT} proposes an unsupervised instance segmentation method that uses a pre-trained mask generation network to generate object proposals, next samples mask triplets from the proposals to capture the hierarchical relations and learn the embeddings. 

Considering the advantages of the Gumbel distribution in modeling extreme events with very low probabilities, the authors in \cite{GOL} propose \textit{GOL} that utilizes the Gumbel activation function for long-tailed object detection and segmentation. Through large-scale experiments, they demonstrate the superiority of Gumbel against Sigmoid and Softmax in detecting/segmenting rare-class objects.

\subsubsection{Loss Reweighting} \label{subsec:lossreweight}

After obtaining the loss on each class or sample, loss reweighting approaches aim at adjusting the importance of their loss values through the weighting factors. Existing approaches along this line can be divided into class-wise reweighting and instance-wise reweighting methods. In Table \ref{tab:Loss7}, we report the formulas of representative methods in this sub-category.

\begin{figure}[h!]
\centering
\includegraphics[page=1,width=1.0\columnwidth]{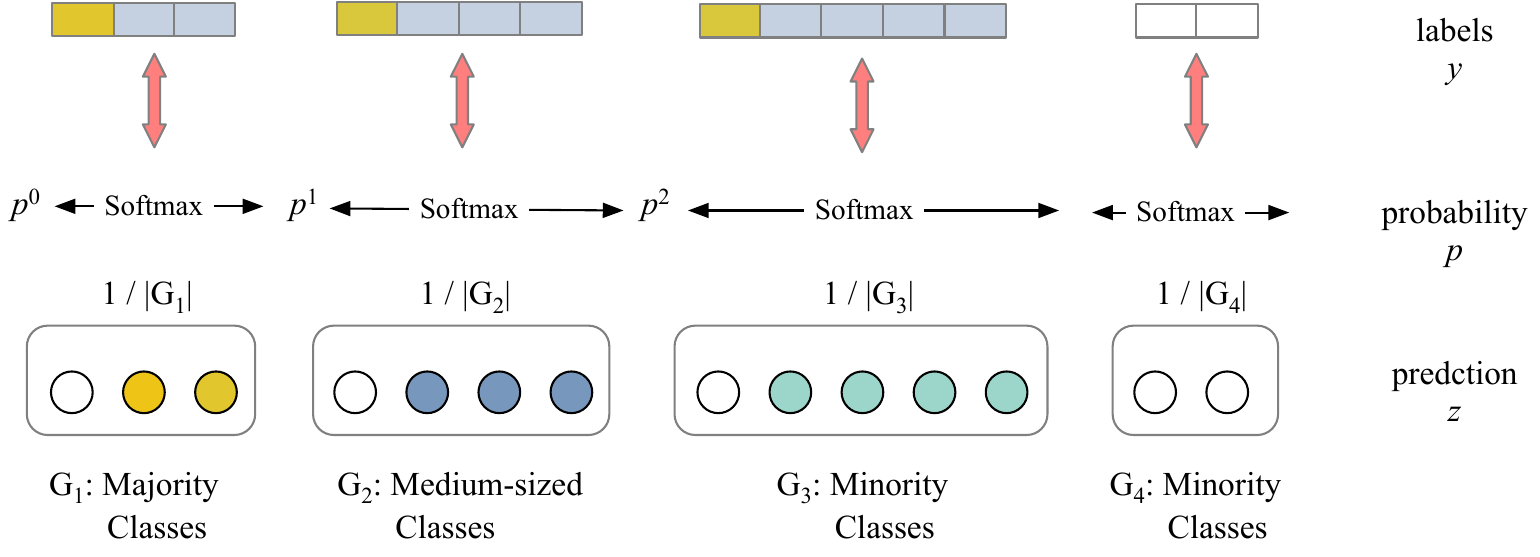}
\caption{Loss reweighting approaches to long-tailed learning.}
\label{fig:lossreweight}
\end{figure}

\textbf{Class-wise Loss Reweighting}. Class-wise loss reweighting methods assign the same weighting factor to instances of the same class or group, as demonstrated in Figure \ref{fig:lossreweight}. CB Loss \cite{cbloss} is a class-balanced loss function for long-tailed distributions that reweights the samples by the inverse class frequency, i.e. the weight/factor of a class is inversely proportional to the number of samples in the class. AREA \cite{AREA} proposes to re-weight the loss of each class based on the inverse Pearson correlation of the samples in the same class.  Range Loss \cite{rangeloss} is also a class-wise reweighting method that consists of an intra-class loss and an inter-class loss: the former is defined as the harmonic mean of the two largest intra-class distances, while the latter is the shortest distance between any two class centres. The Learning to Reweight method (L2RW) \cite{L2RW} is a meta-learning algorithm that assigns weights to a batch of samples based on their similarities in gradient directions.  TCR \cite{balancerethinking} is a two-component reweighting loss function that simultaneously computes the class-wise weights using the CB Loss \cite{cbloss} and the conditional distribution between the source and target with L2RW \cite{L2RW}. 

\textbf{Instance-wise Loss Reweighting}. The above methods assign the same weight to all samples belonging to the same class. In contrast, instance-wise reweighting methods assign different weights to different samples based on their learning difficulty. Focal Loss \cite{focalloss} proposes to assign relatively larger weights to samples with high prediction loss, independent of their class labels. In \cite{EFL}, the authors propose the Equalized Focal Loss (EFL) for long-tailed object detection, which modifies the original Focal loss by supplementing the focusing parameter with a new term that is the inverted ratio between the accumulated positive and negative gradients of the input sample's ground-truth class. In \cite{PSI}, the authors propose the Population Stability Index (PSI) to measure the overall trend that an instance experiences loss increase over time. It pays more attention to difficult instances, including those from the majority classes.  In IB Loss \cite{IBLoss},  the authors propose a method that reweights each sample differently using the designed influence function to alleviate overfitting in the decision boundary. The authors in \cite{MiSLAS} discover that models trained on long-tailed datasets are more miscalibrated and over-confident than those trained on balanced data. They thus propose the label-aware smoothing soft label method  (MiSLAS) that assigns different weights to the head and tail classes to handle different degrees of over-confidence. 

\subsection{Bells and Whistles}
In the context of long-tailed learning, ``bells and whistles'' refer to various techniques, training strategies and practices that go beyond the neural architectures and loss functions but are also essential for improving model performance. Common ``bells and whistles'' include network training and  optimization strategies and choices, such as advanced optimizer, learning rate schedulers, regularization, and commonly adopted techniques in deep learning such as data augmentation, pre-training and transfer learning, knowledge distillation, and ensemble learning for improving the performance of deep learning models. 

There are different types of optimization algorithms used in neural networks \cite{nnoptim}, among them,  Stochastic Gradient Descent (SGD) is the major algorithm for optimizing the parameters (weights and biases) in neural networks for long-tailed learning. Learning rate (LR) schedulers are methods that adjust the learning rate during network training, such as linear decay, step decay (such as CosineAnnealing), time-based decay and exponential decay to optimize model performance over time. Regularization techniques in deep learning are methods used to prevent overfitting and improve the generalization ability of neural networks.  Commonly used regularization techniques include L2 regularization (Weight Decay) \cite{WD,eWD2024}, dropout, batch normalization, label smoothing (which replaces hard labels with smoothed distributions to reduce the model's confidence and encourage it to be more robust and generalize better) \cite{MiSLAS}, etc. 

Since data augmentation and synthesis, knowledge distillation, ensemble (multi-expert) learning methods have already been addressed in the previous subsections,  we only present a few pre-training and transfer learning techniques, and foundation models based approaches. 

\textbf{Pre-training and transfer learning for long-tailed learning}. SSP \cite{ssp} introduces self-supervised pre-training into class-imbalanced learning to yield good network initialisation, which can alleviate the label bias issue in imbalanced datasets and enhance LTL performance. FBGL \cite{FBGL} uses both contrastive pretraining and normalization for improving the feature extractor and classifier for long-tailed recognition, which consists of a new balanced contrastive loss to improve  long-tailed pretraining and a generalized normalization method for the classifier (feature and weight vectors). The authors in \cite{Dong_2023_ICCV} propose a simple yet effective step-wise learning framework based on Deformable DETR \cite{DETR} that combines fine-tuning and knowledge transfer for long-tailed object detection. In specific, after selecting representative examplars for the head and tails classes,  it first pre-trains a model on all the classes, next learns a head-class model via fine-tuning, then transfers knowledge  from the head-class model to the final model. 

\textbf{Leveraging large foundation models and multi-modal models for long-tailed learning}. Lately, there has been a popular trend in utilizing the powerful feature representation capability of foundation models \cite{FL1,FL2,FL3,FV1} and pre-trained multi-modal models \cite{CLIP,VL-LTR}  to improve the LTL performance, which can also be considered as a ``trick'' for long-tailed learning. For instance, integration of textual and visual features  enhances the model's ability to discriminate between classes with fewer examples \cite{VL-LTR,RichSem}, or enables the model to understand and reason about the semantic relationships between concepts depicted in images and described in texts \cite{LPT}.

Finally, we note that, these strategies and techniques are often collectively used to make LTL models more effective and efficient. 

\subsection{Network Optimization}
Network optimization refers to the internal mechanism for updating and optimizing the neural networks to  minimize the loss/objective function. Besides neural architecture and loss function design, optimizing the huge amount of internal learnable parameters of a neural network, mainly the weights and biases, plays a very important role in efficiently and effectively training and optimizing a deep learning model. In particular, this process involves techniques in updating the weights, and the gradients with respect to the loss function and weights.  Existing approaches in this line of research can be divided into weights rebalancing and gradients rebalancing approaches.

We clarify that, ``network optimization'' refers to techniques that optimize or utilize the weights and gradients of the network for internal network optimization, while ``Bells and Whistles'' in the above subsection refer to different external training strategies, choices, techniques and tricks for obtaining better learning performance.

\subsubsection{Weights Rebalancing}
The authors in \cite{kim2020adjusting} show an obvious correlation between sampling frequency and the norm of the weight vectors, and propose to raise the weight vector of each class by multiplying the ratio between the number of samples in the most frequent class and the current class to adjust the decision boundary in long-tailed recognition. The authors in \cite{WD} also reveal that, for long-tailed data, a normally trained deep learning model has imbalanced norms of classifier weights, with the majority classes having dominating weights. They propose to balance network weights using standard regularization techniques such as weight decay (L2 regularization) and notably improve the long-tailed recognition performance. \cite{eWD2024} discusses why weight balancing \cite{WD} is effective in two-stage long-tailed learning. It reveals that  weight decay helps achieve better feature extraction, while  class-balanced loss enables implicit logit adjustment in the training. Curvature-Balanced  Regularizer (CR) \cite{CR} uses the sum of the logarithm of the inverse of the maximum normalized Gaussian curvature mean of each class as a regularizer to enable the model to learn curvature-balanced manifolds. 

\subsubsection{Gradients Rebalancing}
\cite{10143656} reveals that long-tailed data suffers from gradient distortion problem in which the overall gradient is shifted toward the head gradient,  it thus proposes a two-phase long-tailed recognition method which first updates model parameters  using the gradient on  the head classes only, next grows the classifier on the added tail classes. The authors in \cite{BSGAL} propose BSGAL which is a gradient-based generated data contribution estimation method for long-tailed instance segmentation that calculates the loss and gradient of the model on a batch of data with and without augmentation to actively accept or reject the generated data. SeeSaw Loss \cite{SeeSaw} dynamically rebalances the gradients of positive and negative samples for each category with two complementary factors to reduce punishments on tail categories and increase the penalty of misclassified instances. For long-tailed class-incremental learning, \cite{He_CIL} proposes to separately reweight the gradients for the new tasks and learned tasks, based on the accumulated magnitude of gradients and the volume of the lost training data of each class, respectively. 

Overall, properly tuning the above factors is crucial for the internal optimization of the neural network to achieve high model accuracy and prevent overfitting in long-tailed learning.

\subsection{Post Hoc Processing}
\begin{figure}[h!]
\centering
\resizebox{1.0\columnwidth}{!}
{
\includegraphics[page=1,width=0.9\columnwidth]{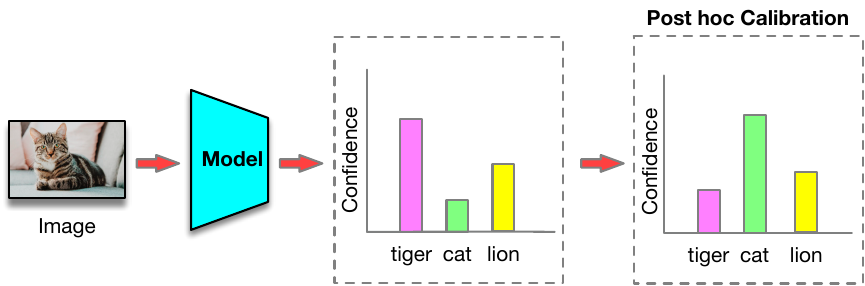}
}
\caption{Post hoc processing methods for long-tailed learning.}
\label{fig:posthoc}
\end{figure}

As demonstrated in Figure \ref{fig:posthoc}, post hoc processing methods modify the outputs of the (conventionally) trained models to tackle the imbalance problem at the inference stage, instead of improving the feature representation capability or adjusting the decision boundary at the model training process. The expected calibration error (ECE) \cite{ECE} is often used to measure how well the predicted probabilities match the ground-truth probabilities, which first partitions the confidence scores into several equal-size bins,  then computes the difference between the accuracy and the averaged confidence in each bin.

\subsubsection{Post Hoc Confidence Calibration}
Temperature Scaling (TS) \cite{TS} uses a single scalar parameter T to rescale the logits of a trained neural network, which will then be transformed into calibrated confidence scores using the Softmax activation function. However, this approach has yet to be validated in long-tailed settings. 

For confidence calibration in long-tailed object detection, NORCAL \cite{NORCAL}  multiplies the prediction (output) scores with a vector of factors that are inversely proportional to the training sample size of each class. 

For confidence calibration in long-tailed object segmentation, the Pairwise Class Balance (PCB) loss \cite{PCB} post-calibrates a sample's prediction probability on each class by multiplying the corresponding pairwise misclassification bias obtained from the confusion matrix. 

\subsubsection{Confidence Aggregation}
When a neural network has multiple heads for inference, e.g. the dual streams inference framework and the multi-expert learning framework addressed in subsections \ref{dualinfer} and \ref{multiexpert}, these methods commonly use the logits mean \cite{RIDE} or the weighted aggregation of the logits \cite{SADE}, as well as the confidence mean to make the final prediction.

\section{Comparisons Between Long-Tailed Learning and Imbalance Learning} \label{sec:imltlcom}

Imbalance learning is a long-standing research direction in machine learning \cite{hesurvey, krawczykimsurvey, cusrsurvey}.  Long-tailed learning  belongs to imbalance learning, but they also have differences in the learning approaches. While conventional imbalance learning algorithms commonly deal with rational/tabular imbalanced data, long-tailed learning often handles image data, which needs the feature representation learning step as a prerequisite. That is, for image data, LTL needs deep representation learning techniques to extract features from images, in which neural architecture, feature enrichment and logit adjustment techniques are needed. 

However, for relational/tabular data, its features are its attribute  values, therefore,  features are already available before imbalance learning, thus feature representation techniques such as neural architecture, feature enrichment and logits adjustment, are not essential in imbalance learning. But we note that, in recent years,  deep imbalance learning approaches have also emerged, which use neural networks to improve the imbalance learning performance on relational/tabular data. 

In the following, we will mainly compare the data balancing and loss functions/cost-sensitive learning methods, as well as the multi-expert/ensemble learning methods adopted in both LTL and imbalance learning. 

\subsection{Data Balancing}

\subsubsection{Data Resampling}

In long-tailed learning, only random sampling is utilized, either at class-level or instance-level, referred to as class-wise sampling and instance-wise sampling, respectively. However, in imbalance learning, more advanced undersampling and oversampling techniques have been proposed. 

\textbf{Undersampling techniques in imbalance learning}. In Condensed Nearest Neighbor Rule \cite{imcnn}, for each sample from the majority set, if it can not be correctly classified by the Nearest Neighbor (NN) rule, it is kept; otherwise, it is discarded. In \cite{zhu2018geometric}, the authors propose an intuitive geometric space partition (GSE) based method for imbalance classification which iteratively uses a new hyper-plane classifier to cut the current geometric data space into two partitions, then removes the partition that only contains the majority of samples. 

\textbf{Oversampling techniques in imbalance learning}. ROS  \cite{ros} is the most straightforward oversampling technique, which  balances the data by replicating the minority class samples. The Majority Weighted Minority Oversampling Technique (MWMOTE) \cite{MWMOTE} first identifies the most important and hard-to-learn minority class samples, then assigns weights to them according to their Euclidean distance to the nearest majority class samples.

\subsubsection{Data Augmentation}

For image data, in long-tailed learning, Mixup \cite{mixup}, ReMix \cite{Remix} and UniMix \cite{UniMix} create synthetic samples with interpolated feature vectors and labels out of two original samples.

For relational/tabular data, since the features are already available before imbalance learning, it is easy and common to use linear interpolation algorithms to generate synthetic minority samples between two real minority samples of the same class. SMOTE \cite{smote} creates synthetic minority samples by randomly selecting a new point on the line segment between a pair of neighbouring minority samples. Borderline SMOTE \cite{borderlinesmote} extends SMOTE by emphasizing the borderline samples that have both majority and minority class points in their neighbourhood and ignoring the rest of the minority samples. ADASYN \cite{adasyn} creates synthetic minority samples according to data density, where a comparatively larger number of synthetic samples are created (using SMOTE) in regions of low density of the minority class than in higher-density regions. SMOTE-RSB \cite{SMOTE-RSB} first uses SMOTE to generate synthetic instances for the minority classes, then employs the rough set theory as a cleaning (undersampling) method to select only the synthetic minority examples that belong to the lower approximation of their class.

In general, the idea of Mixup, ReMix and UniMix is similar to SMOTE  and its variants, but the former interpolates the label space as well, while SMOTE and its variants only handle samples of the same minority classes,  but without modifying the labels of the synthetic samples. 

\subsubsection{Data Synthesis}

In LTL, there are distribution based data synthesis methods, such as (RSG) \cite{RSG}, LEAP \cite{LEAP} and GistNet \cite{GistNet}, GAN based \cite{UTLO} and foundation model based data generation \cite{LTGC} methods. In comparison, in imbalance learning,  GANs are commonly utilized to generate synthetic tabular data, such as TableGAN \cite{TableGAN}, CTGAN \cite{CTGAN}, medGAN \cite{medgan} and QAST \cite{qastaaai2023}. 

TableGAN \cite{TableGAN} was the first attempt to synthesize tables/relational databases using GANs. It adopts the Deep Convolutional Generative Adversarial Network (DCGAN) architecture, but adds a new neural network module called ``classifier'' and a corresponding loss function to measure the discrepancy between the label of a generated record and the label predicted by the classifier for this record. CTGAN \cite{CTGAN} models the distribution of the values with a few Gaussian distributions (modes) for each continuous attribute. For discrete attributes, CTGAN introduces a conditional generator and training-by-sampling method to ensure that each category of a discrete attribute gets a fair chance to be included during sampling. This way, it can simultaneously generate a mix of discrete and continuous attributes. Medical Generative Adversarial Network (medGAN) \cite{medgan} uses Auto-Encoders to learn the salient features of discrete attributes and GAN to generate realistic samples in the embedding space. To address the usability of GAN-generated samples in imbalanced learning,  QAST \cite{qastaaai2023} designs a semantic pseudo-labelling module to control the quality of the generated features, then calibrates their corresponding semantic labels using a classifier committee consisting of multiple pre-trained shallow classifiers. The above approaches only generate synthetic examples for the minority classes. In \cite{jing2021multiset}, the authors propose using GAN to generate a subset of the majority of samples that follow the same distribution as the original dataset. Next, they construct a few balanced subsets, each containing all the minority samples and an equal-sized subset of GAN-generated majority samples. Then, upon each decomposed two-class subset, they use deep metric/embedding learning to build a two-class classification model. These models are then integrated to make the final predictions.  Recently, a diffusion model \cite{diffusion} based imbalance learning method, namely TabDDPM  \cite{TabDDPM},  has been proposed for generating rational samples.

\subsection{Loss Function}

\subsubsection{Loss Function for Embedding Learning}

The idea behind embedding learning is to map complicated (often high-dimensional) metric spaces into easier (often low-dimensional) metric spaces, where inter-class and intra-class distances can be optimized simultaneously. In LTL, there are many specially designed embedding learning approaches, such as \cite{klimb}, \cite{DLSA}, \cite{CReST}, \cite{kcl}, \cite{BCL}, etc.

In imbalance learning over relational data, many traditional (non-deep learning based) embedding learning methods have been proposed to transform the learning space. IML \cite{IML} is an imbalance learning method that aims to find a more stable neighbourhood space for the testing data by performing metric learning on the data and training sample selection according to the testing data iteratively until the selected training samples in two adjacent iterations are the same. MLFP \cite{MLFP}  expands the decision boundaries around the positive samples and learns better metric space using a triplet-based cost function on the false negative and false positive samples. 

In recent years, many deep imbalance learning approaches have also been proposed which use neural networks to transform the metric space and learn better embedding/feature representations. Khan et al. \cite{imcost2018} propose a cost-sensitive (CoSen) deep neural network method to learn feature representations for both the majority and minority classes. \cite{imrect2019} addresses the problem of imbalanced data for the multi-label classification problem using deep learning. It introduces incremental minority class discrimination learning by formulating a class rectification loss regularization, which imposes an additional batch-wise class balancing on top of the cross-entropy loss. This can rectify model learning bias due to the over-representation of the majority classes. GAMO \cite{gamo} is an end-to-end deep oversampling model for imbalance classification, which effectively integrates sample generation with classifier training. 

\subsubsection{Loss Reweighting}
In LTL, loss-level reweighting methods assign different weights to the loss values of the samples, which include class-wise reweighting algorithms such as CB Loss \cite{cbloss} and L2RW \cite{L2RW}, instance-wise reweighting algorithms such as IB Loss \cite{IBLoss}, as illustrated in subsection \ref{subsec:lossreweight}. 

In imbalance learning, there are also a few cost-sensitive learning approaches that modify the internal cost function/objective function by assigning different weights to different samples based on their difficulty. Such methods usually assign a higher cost to minority classes to boost the importance of the minority classes during the learning process \cite{zhou2010multi}. To adapt SVM to imbalanced data, cost-sensitive SVM (CostSVM) \cite{CostSVM} adds a class-specific classification penalty in its cost function. 

Overall, the general ideas for loss-level reweighting in LTL and cost-sensitive learning in imbalance learning are very similar, i.e., they all give higher weights to samples of the minority classes or the misclassified samples. 

\subsection{Multi-Expert Learning}

Multi-expert learning (ensemble learning) techniques are applicable to both LTL and imbalance learning.  For instance, the idea of ACE \cite{ACE} to train cascaded learning models for LTL is similar to that of BalanceCascade \cite{easyensemble} and AdaCost \cite{adacost} for imbalance learning. 

There are also a few differences, for instance, i) Knowledge distillation-based ensemble techniques have been adopted in LTL \cite{SADE,LFME,CBD21}, but not in imbalance learning; ii) Dichotomy-based ensemble technique for multi-class imbalance learning \cite{decoc,oaho} are not needed in long-tailed learning, since cross-entropy loss based LTL classifiers are naturally multi-class classifiers.

\subsection{Evaluation Metrics}

In terms of evaluation metrics,  existing LTL methods commonly assume that the test sets are class-balanced, that is, all the classes have the same number of test instances in the testing/inference stage. Under such an assumption, only the Top-1 accuracy/overall accuracy metric is reported. However, we note that, in real-world applications, the test sets might also be long-tailed, or at least non-uniformly distributed.  In contrast, in imbalance learning the test sets are often considered to be imbalanced, and various evaluation metrics have been proposed \cite{hesurvey,raeder2012learning}, including the overall accuracy, G-mean (the geometric mean of the per-class accuracy), Macro-F1, etc. 

In \cite{GML}, the authors argue that the average accuracy metric used in LTL incurs little penalty on classes with very low per-class recalls. They thus propose three new metrics for evaluating long-tailed recognition algorithms: the geometric mean and harmonic mean of the per-class recall (referred to as G-mean and H-mean for short), and the lowest per-class recall. They also propose the Geometric Mean Loss (GML), which essentially calculates the logarithm of the averaged prediction probability for samples of each class. 

As a short summary, when the test set distributions are non-uniform or skewed, major evaluation  metrics including overall accuracy, G-mean,  Macro-F1 and the lowest per-class recall, should  be adopted in LTL and imbalance learning to more comprehensively assess the performance of the proposed algorithms.

\section{Long-Tailed Distributions: a Discussion} \label{sec:ltdist}

\subsection{Comparing Long-tailed Data with Imbalanced  Data}

When referring to long-tailed data, one may neglect the fact that long-tailed data is a special type of imbalanced data, and long-tailed learning (LTL)  is essentially part of imbalance learning (IL). 

The long tail phenomenon occurs in imbalanced data when a small number of classes are highly frequent and are thus easy to model. In the meantime, there are also a substantial number of tail classes that have insufficient samples and are thus difficult to model. Moreover, the combined quantity or influence of all the tail classes is comparable to that of the head classes. Specifically, there are three main differences between long-tailed distributed data and imbalanced data:

First, in long-tailed data, there should be a significantly large number of minority (tail) classes compared to the majority (head) classes. It thus has a very long tail, also named ``fat tail'' or ``heavy tail''. However, in the case of imbalanced data, the number of classes may be as few as two (two-class imbalanced data) or only a few (multi-class imbalanced data). Therefore, long-tailed data must be class-imbalanced data, but class-imbalanced data is not necessarily long-tailed data.
  
Second, since long-tailed data has a very long tail, the overall trend/pattern of event frequency/sample size per class should decrease exponentially from the head classes to the tail classes, but the sub-pattern/trend among the tail classes should be a relatively flat curve with a slowly decreasing pattern. Ordinary class-imbalanced data does not necessarily have this property.
  
Third, the combined quantity or influence of the tail class samples in long-tailed data should be significant, e.g. 60\% or 20\% of the total quantity/influence of the head classes. This unique property does not hold in ordinary class-imbalanced data.

\subsection{Comparing Long-tailed Distribution with Pareto Distribution}

A Pareto distribution is a power-law probability distribution with heavy, slowly decaying tails. In comparison, a ``long-tailed distribution''  essentially corresponds to a set of qualitative properties/constraints added to a class-imbalanced distribution rather than being a statistical distribution that can be rigorously modelled. In the following, we empirically study the differences between a Pareto distribution and a long-tailed distribution.

\begin{figure*}[h!]
\centering
 \begin{subfigure}[b]{\columnwidth}
 \includegraphics[width=0.98\columnwidth]{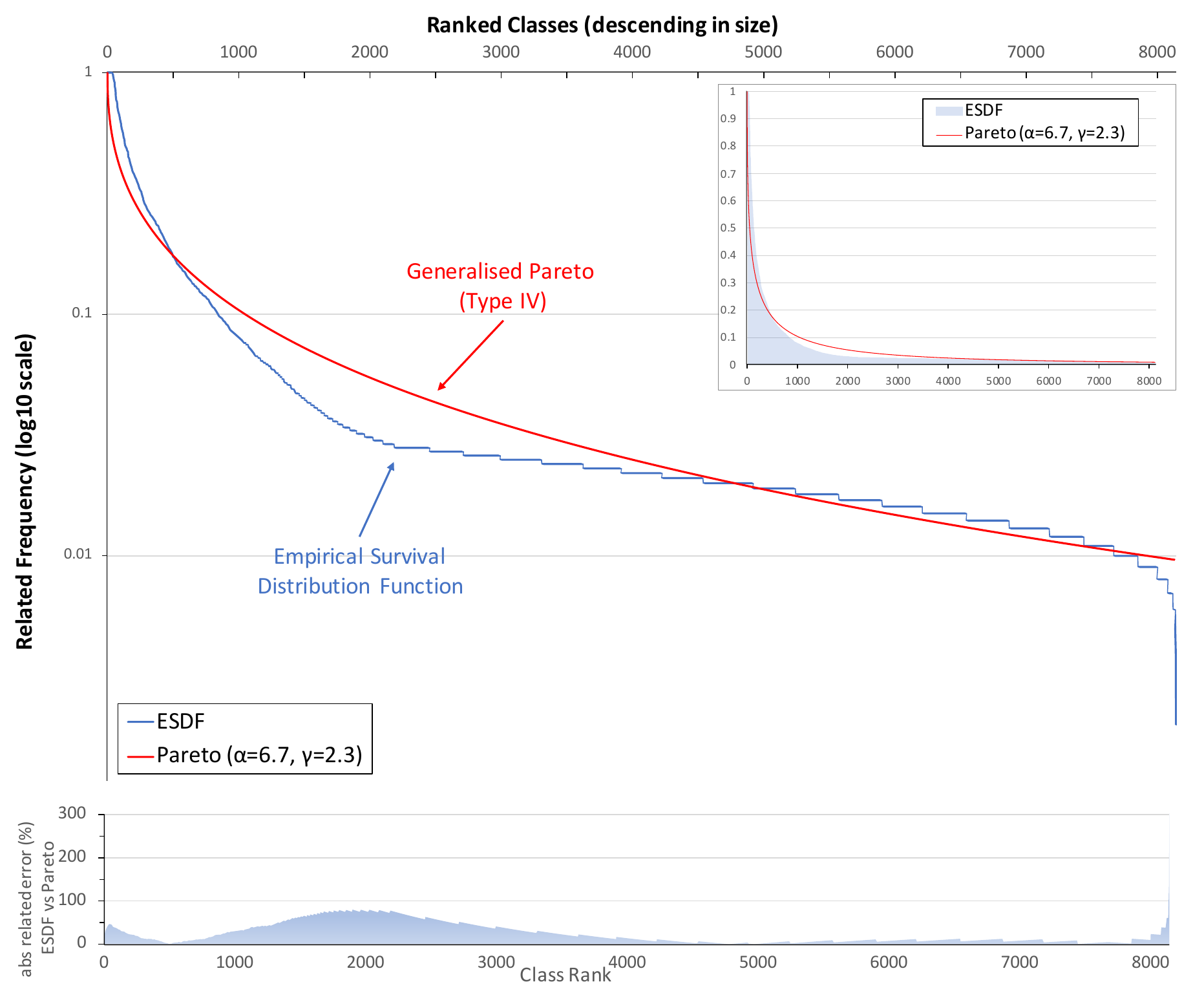}
 \caption{iNaturalist 2018}
 \end{subfigure}
 \begin{subfigure}[b]{\columnwidth}
 \includegraphics[width=0.98\columnwidth]{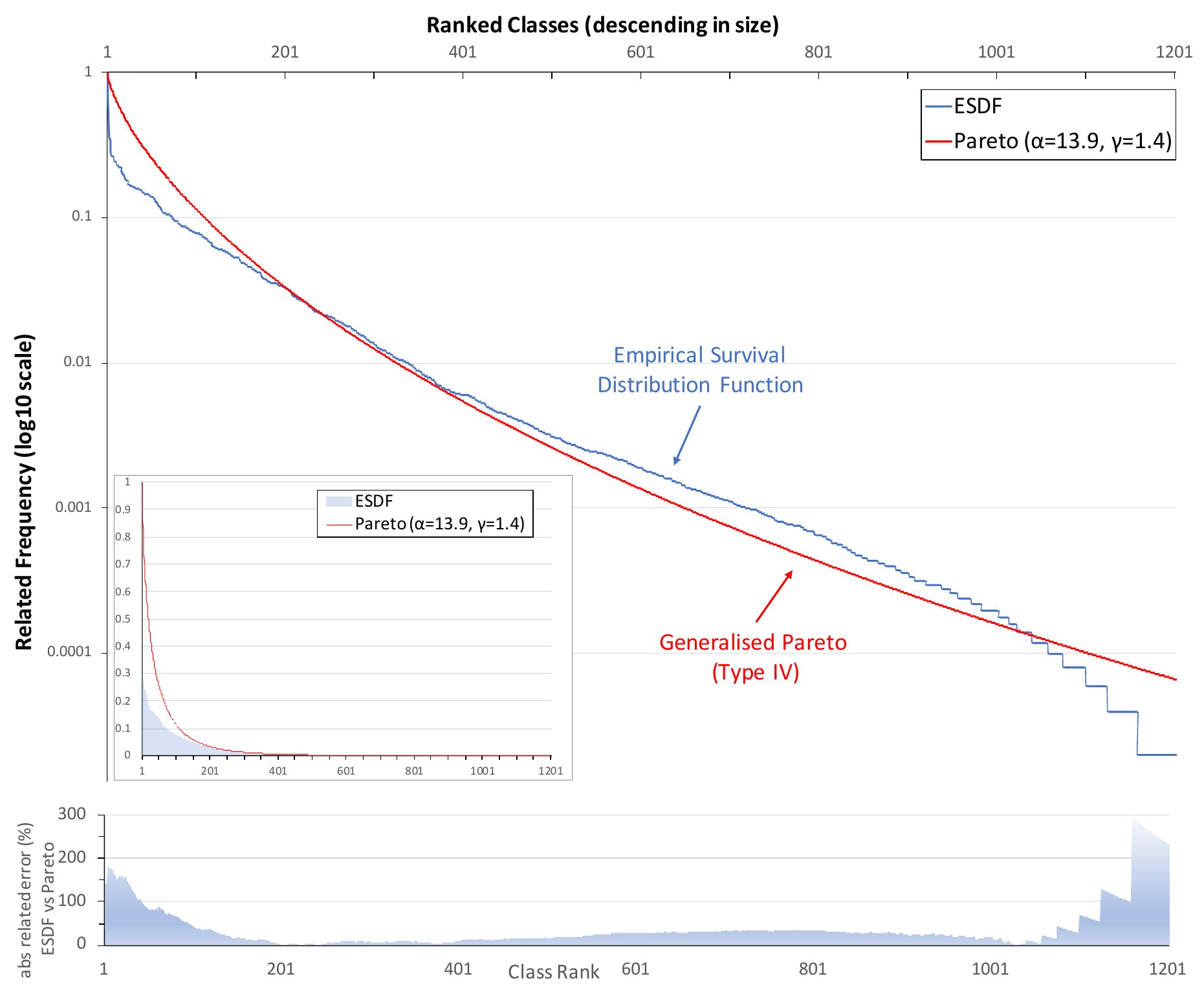}
 \caption{LVIS}
 \end{subfigure}
\caption{Empirical distributions of two real-world long-tailed datasets fitted with Generalised Pareto distributions.}
\label{fig:datasets}
\end{figure*}

In Figure \ref{fig:datasets}, we first depict the empirical rank-size distributions of two real-world datasets, iNaturalist 2018 and LVIS, typically used in long-tailed learning settings. A rank-size distribution gives the empirical relative frequencies  of the data with the classes ordered in descending order in size across the x-axis\footnote{When calculating the relative frequencies, for normalisation reasons, we divide the sample size/frequency of each class by the maximum frequency (the frequency of the Top-1 class), instead of the total number of samples of all the classes.}, which corresponds to the empirical survival distribution function (ESDF).

\begin{equation}
\label{eq:dpivsf}
S(x)=\left[ 1 + \left( \frac{x-\mu}{\sigma} \right) ^ {\frac{1}{\gamma}} \right] ^ {-\alpha} , x\geq\mu , \mu \in \mathbb{R} , \sigma , \gamma > 0, \alpha \geq 1
\end{equation}

For comparing a long-tailed distribution with a Pareto distribution, in Figure \ref{fig:datasets} we fit each dataset with a Generalised Pareto Type IV (DPIV) survival function (SF), using Equation \ref{eq:dpivsf}, where  $\mu$ is the location, $\sigma$ is the scale (which corresponds to the minimum value of $x$),  $\gamma$ is the inequality parameter, and $\alpha$ is the tail index (shape). Robust estimators for the parameters of DPIV have been proposed in the literature. Here, after fixing parameters $\mu$ and $\sigma$ in relation to the empirical distribution of each dataset, we exhaustively vary the values of $\alpha$ and $\gamma$ with a step of $s=0.1$ to heuristically find the set of parameter values that minimises the difference between the ESDF and the fitting Pareto survival function, measured in terms of the relative errors/distances between them, which can be observed from the bottom sub-figures of Figure \ref{fig:datasets}.

As the Pareto distribution (DP) embodies a heavy (fat) tail, it is natural to present the survival rates of both the theoretical and the empirical distribution on a logarithmic scale. In Figure \ref{fig:datasets}, on the two real-world long-tailed datasets iNaturalist and LVIS, we see that the derived Pareto distribution is not guaranteed to consistently fit to all segments of the ESDF of the real long-tailed data.

In summary,  unlike the Pareto distribution, a long-tailed distribution is not a specific statistical distribution that can be precisely modelled, but rather an imbalanced distribution with a set of properties/constraints.

\subsection{When Using the Pareto Distribution to Model the Long Tail Phenomena}

In case a Pareto distribution is used to approximately model the long tail phenomena, we need to add the following constraint to the Pareto distribution:

\textit{In long-tailed data, the combined quantity/significance of the tail class samples should exceed a user-specified ratio compared to the head classes, e.g. 60\%, or 50\%, or 20\%.}

The Cumulative Distribution Function (CDF) of the Pareto Distribution is given by:

\begin{equation}
 F(x)  = 1 - (\frac{x_m}{x})^\alpha
\end{equation}

where $x_m$ is the scale parameter, e.g. $x_m$ = 10 or 3, which can be considered as the minimum number of samples that a tail class has.  $\alpha$ is the shape parameter, which is essentially the power law slope. 

Based on the Pareto Distribution, the constraint added by the long-tailed data depicted above can be formulated as: 

\begin{equation}
\bar{F}(h_l)= P(X \geq h_l) = (\frac{x_m}{h_l})^\alpha \leq 1 - r
\label{eq4}
\end{equation}

where $h_l$ is the lower bound frequency of the head classes, i.e. the number of samples of the least-ranked head class, and $r$ is a user-specified threshold which is the ratio between the combined quantity/influence of the tail classes and the total quantity/influence of all the classes.  Different $r$ values can be set in different circumstances; for instance, $r$ can be 60\%, 50\%, or 20\%.  Essentially, for modeling long-tailed data using the Pareto distribution, Equation \ref{eq4} adds to the Pareto distribution a joint constraint on the parameters $x_m$, $\alpha$, $h_l$ and $r$.

\section{Performance Analysis} \label{sec:perform}

\subsection{Datasets}
\label{sec:datasets}

For long-tailed recognition, benchmark datasets include CIFAR-100-LT \cite{marginimloss}, ImageNet-LT \cite{openlongtail}, Places-LT \cite{openlongtail}, and iNaturalist 2018 \cite{iNaturalist}. For long-tailed object detection and instance segmentation, LVIS \cite{LVIS} is the primary real-world dataset. LVIS v0.5 consists of 1,230 categories, whereas LVIS v1.0 contains 1,203 categories, with 100K images and 19.8K images for the training and test sets, respectively. 

\subsection{Evaluation Metrics}
For long-tailed recognition tasks, since existing LTL methods all assume that the test set is uniformly distributed, i.e.,  the sample size for each class in the test set is equal, therefore only the overall accuracy metric is needed for performance evaluation. Besides, these methods often separately report the specific performance on three super groups, which are ``Many-shot'' (if a class has > 100 instances), ``Medium-shot'' (20-100 instances) and ``Few-shot'' (< 20 instances).

For long-tailed detection and segmentation tasks,  the main evaluation metric is Average Precision (AP). For long-tailed segmentation, separate performance on three super groups are also reported, which are ${AP}_{f}$ for frequent classes (having > 100 instances), ${AP}_{c}$ for  common class (having 11-100 instances), and ${AP}_{r}$ for rare classes (having < 11 instances).

\subsection{Performance Comparison}

\subsubsection{Long-tailed Recognition}

Table \ref{tab:cls} summarizes the performance of representative long-tailed classification algorithms on four benchmark datasets CIFAR-100-LT (with two different imbalance ratios (IR) which are 50 and 100), ImageNet-LT,  Places-LT, and iNaturalist 2018. It can be observed that, under the CNN backbone (ResNet), MDCS \cite{MDCS}, LGLA \cite{LGLA} and SHIKE \cite{SHIKE} achieve the best overall performance on all the four benchmark datasets. Besides, NCL \cite{NCL}, SADE \cite{SADE},  RIDE \cite{RIDE} and ACE \cite{ACE} also obtain remarkable performance, all of which adopt the multi-expert learning framework, owing to its effectiveness. While under the Transformer backbone \cite{ViT}, due to the utilization of foundation models, LTGC \cite{LTGC} and VL-LTR \cite{VL-LTR} achieve significantly better performance than CNN based methods, in particular, their performance improvements over MDCS \cite{MDCS} and LGLA \cite{LGLA} on ImageNet-LT  are near 20\% in terms of overall accuracy, which is very substantial. 

Besides, we see that the performance of loss reweighting methods tend to saturate, since almost all the possible forms have been exhaustively examined, as can be observed from Table \ref{tab:Loss7}. 

In Table \ref{tab:expsetting}, we report the experimental settings of representative long-tailed recognition algorithms on the iNaturalist 2018 dataset. 

Overall, on ImageNet-LT and iNaturalist 2018, the state-of-the-art performance is only 82.5\% in terms of overall accuracy, while the best overall performance is only 54.1\% on Places-LT, which can hardly satisfy the requirements of real-world applications. Therefore, there is still a large room for performance improvement.

\subsubsection{Long-tailed Object Detection and Segmentation}
Table \ref{tab:seg_dec} reports the performance of representative long-tailed object detection and segmentation methods on the LVIS v1.0 benchmark dataset. We see that  BSGAL \cite{BSGAL}, RichSem \cite{RichSem}, Step-wise \cite{Dong_2023_ICCV}, PCB \cite{PCB}, ECM \cite{ECM}, C2AM \cite{C2AM} achieve the best long-tailed object detection performance, while BSGAL \cite{BSGAL},  GOL \cite{GOL}, ECM \cite{ECM}, PCB \cite{PCB}, C2AM \cite{C2AM} and NORCAL \cite{NORCAL} are among the best performers in long-tailed segmentation. We would like to note that, Step-wise \cite{Dong_2023_ICCV} is implemented based upon Deformable DETR \cite{DETR} which is essentially a CNN-Transformer combined detection framework, while the other methods are commonly based on Mask R-CNN with ResNet as the backbone. 

Overall, the state-of-the-art performance is below 35.4\% in both  long-tailed object detection and segmentation tasks, and the segmentation performance on rare classes is even lower, which is less than 25.4\%. Therefore, there is considerable room for future research.

\begin{table*}[!t]

\centering
\caption{Performance of representative long-tailed recognition methods on the benchmark datasets. A, D, M,  L and W are short for neural architecture and training strategy tuning, data-level rebalancing, multi-experts learning (which is a sub-category under neural architecture and training strategy tuning), logits calibration and margin adjustment,  loss reweighting, respectively. }
\label{tab:cls}
\setlength{\tabcolsep}{0.8mm}
\renewcommand{\arraystretch}{1.5}
{
\begin{tabular}{c|c|c|ccc|ccccc|cc|cc|c}
\bottomrule
\multirow{2}{*}{\moren{\ubold{Method}}} & \multirow{2}{*}{\moren{\ubold{Year}}} & \multirow{2}{*}{\moren{\ubold{Venue}}} &
\multicolumn{3}{c|}{\ubold{CIFAR-100-LT}} &
\multicolumn{5}{c|}{\ubold{ImageNet-LT}} &  
\multicolumn{2}{c|}{\ubold{Places-LT}} & 
\multicolumn{2}{c|}{\ubold{iNaturalist 2018}} & 
\multirow{2}{*}{\moren{
\makecell[c]{\ubold{Type}}
}} 
\\

\cline{4-15}
& & & {\ubold{Epoch}} & {\ubold{IR=100}} & {\ubold{IR=50}} &{\ubold{Epoch}} & {\ubold{Many}} & {\ubold{Medium}} & {\ubold{Few}} & {\ubold{Overall}} & 
\ubold{Epoch} & {\ubold{Overall}} & \ubold{Epoch} & {\ubold{Overall}} \\
\hline

\multicolumn{3}{l}{\ubold{\textit{CNN Backbone (ResNet)}}}
&\multicolumn{3}{c}{\ubold{ResNet-32}} &
\multicolumn{5}{c}{\ubold{ResNet-50}}
& \multicolumn{2}{c}{\ubold{ResNet-152}}
& \multicolumn{2}{c}{\ubold{ResNet-50}}
\\
\hline

\multicolumn{1}{l|}{TCR Loss \cite{balancerethinking}}& 2020 & \it CVPR 
& 200 & 44.1 & 49.2
& 90 & - & - & - & 48.0 
& 30 & 30.8
& 90 & 67.6 & W \\

\multicolumn{1}{l|}{cRT \cite{decouple}}& 2020 & \it ICLR 
& 200 & 43.8 & -
& 100 & 58.8 & 44.0 & 26.1 & 47.3 
& 30 & 36.7
& 100 & 62.5 & A \\

\multicolumn{1}{l|}{BBN \cite{BBN}}& 2020 & \it CVPR 
& 200 & 42.6 & 47.0
& - & - & - & - & - 
& 30 & - 
& 200 & 69.6 & A\\

\multicolumn{1}{l|}{LFME \cite{LFME}}& 2020 & \it ECCV 
& 200 & 43.8 & -
& -& - & - & - & - 
& 30 & 36.2
& 90 & 69.6 & M \\

\hline

\multicolumn{1}{l|}{MetaSAug \cite{MetaSAug}}& 2021 & \it CVPR 
& 200 & 48.0 & 52.3
& 90 & - & - & - & 47.4 
& 30 & -
& 100 & 68.8 & D \\

\multicolumn{1}{l|}{IB Loss \cite{IBLoss}}& 2021 & \it ICCV 
& 200 & 42.1 & 46.2
& - & - & - & - & - 
& 30 & -
& 90 & 65.4 & W \\

\multicolumn{1}{l|}{LogitAdjust \cite{logitlongtail}}& 2021 & \it ICLR 
& 200 & 44.1 & -
& 90 & - & - & - & 48.8 
& 30 & - 
& 90 & 68.4 & L \\

\multicolumn{1}{l|}{LADE \cite{LADE}}& 2021 & \it CVPR 
& 200 & 45.4 & 50.5
& 90 & 62.3 & 49.3 & 31.2 & 51.9 
& 30 & 38.8
& 90 & 70.0 & L \\

\multicolumn{1}{l|}{DisRobuLT \cite{DisRobuLT}}& 2021 & \it ICCV 
& 200 & 47.3 & 57.6
& 90 & 64.0 & 49.8 & 33.1 & 53.5 
& 30 & - 
& 90 & 69.7 & L \\

\multicolumn{1}{l|}{DisAlign \cite{DisAlign}}& 2021 & \it CVPR 
& - & - & -
& 90 & 61.3 & 52.2 & 31.4 & 52.9 
& 30 & 39.3
& 90 & 70.2 & A \\

\multicolumn{1}{l|}{DiVE \cite{DiVE}}& 2021 & \it ICCV 
& 200 & 45.4 & 51.1
& 90 & 64.1 & 50.4 & 31.5 & 53.1 
& 30 & -
& 90 & 71.7 & A \\

\multicolumn{1}{l|}{SSD \cite{SSDLT}}& 2021 & \it ICCV 
& 200 & 46.0 & 50.5
& 135 & 66.8 & 53.1 & 35.4 & 56.0 
& 30 & -
& 90 & 71.5 & A \\

\multicolumn{1}{l|}{PaCo \cite{PaCo} }& 2021 & \it ICCV 
& 400 & 52.0 & 56.0
& 400 & 65.0 & 55.7 & 38.2 & 57.0 
& 30 & 41.2
& 400 & 73.2 & A \\

\multicolumn{1}{l|}{Hybrid \cite{wangcontrastive}}& 2021 & \it CVPR 
& 200 & 46.7 & 51.9
& - & - & - & - & - 
& 30 & -
& 90 & 68.1 & M \\

\multicolumn{1}{l|}{RIDE \cite{RIDE}}& 2021 & \it ICLR 
& 200 & 48.0 & -
& 100 & 66.2 & 51.7 & 34.9 & 54.9 
& 30 & 40.3
& 100 & 72.2 & M \\

\multicolumn{1}{l|}{ACE \cite{ACE}}& 2021 & \it ICCV 
& 400 & 49.4 & 50.7
& 100 & - & - & - & 54.7 
& 30 & -
& 100 & 72.9 & M \\

\hline

\multicolumn{1}{l|}{SAFA \cite{SAFA}}& 2022 & \it ECCV 
& 200 & 46.0 & 50.0
& - & - & - & - & - 
& 30 & 41.5
& 90 & 69.8 & D \\

\multicolumn{1}{l|}{CMO (+RIDE) \cite{CMO}}& 2022 & \it CVPR 
& 200 & 50.0 & -
& 100 & 66.4 & 53.9 & 35.6 & 56.2 
& 30 & -
& 200 & 72.8 & D \\

\multicolumn{1}{l|}{TSC \cite{TSC}}& 2022 & \it CVPR 
& 200 & 43.8 & 47.4
& 100 & 63.5 & 49.7 & 30.4 & 52.4 
& 30 & -
& 100 & 69.7 & L \\

\multicolumn{1}{l|}{DOC \cite{vMFECCV2022}}& 2022 & \it ECCV 
& 200 & - & -
& 90 & 64.2 & 51.4 & 31.8 & 53.7 
& 30 & -
& 100 & 71.0 & L \\

\multicolumn{1}{l|}{GCL \cite{GCL}}& 2022 & \it CVPR 
& 200 & 48.7 & 53.6
& 100 & - & - & - & 54.9 
& 100 & 40.6
& 100 & 72.0 & L \\

\multicolumn{1}{l|}{DLSA (+RIDE)\cite{DLSA}}& 2022 & \it ECCV 
& 200 & - & -
& - & \ubold{\underline{67.8}} & 54.5 & 38.8 & 57.5 
& 30 & -
& 100 & 72.8 & A \\

\multicolumn{1}{l|}{BCL \cite{BCL}   }& 2022 & \it CVPR 
& 200 & 51.9 & 56.6
& 90 & - & - & - & 56.0 
& 30 & -
& 100 & 71.8 & M \\

\multicolumn{1}{l|}{SADE \cite{SADE}}& 2022 & \it NIPS 
& 200 & 49.4 & -
& 100 & 66.5 & \ubold{\underline{57.0}} & 43.5 & 58.8 
& 30 & 40.9
& 200 & 72.9 & M \\

\multicolumn{1}{l|}{NCL \cite{NCL} }& 2022 & \it CVPR 
& 400 & 54.2 & 58.2
& 400 & - & - & - & 59.5 
& 30 & 41.8
& 400 & 74.9 & M \\

\hline

\multicolumn{1}{l|}{CUDA (+RIDE) \cite{CUDA}}& 2023 & \it ICLR 
& 200 & 50.7 & 53.7
& 100 & 65.9 & 51.7 & 34.9 & 54.7 
& 30 & -
& 100 & 72.4 & D \\

\multicolumn{1}{l|}{AREA \cite{AREA}}& 2023 & \it ICCV 
& 200 & 48.8 & 51.8
& 120 & - & - & - & 49.5 
& 30 & -
& 200 & 68.3 & W \\

\multicolumn{1}{l|}{SBCL \cite{SBCL}}& 2023 & \it ICCV 
& 200 & 44.9 & 48.7
& 90 & 63.8 & 51.3 & 31.2 & 53.4 
& 30 & -
& 100 & 70.8 & L \\

\multicolumn{1}{l|}{CR (+RIDE) \cite{CR}}& 2023 & \it CVPR 
& 200 & 49.8 & 53.7
& - & - & - & - & - 
& 30 & -
& 100 & 73.5 & L \\

\multicolumn{1}{l|}{CC-SAM \cite{CC-SAM}}& 2023 & \it CVPR 
& 200 & 50.8 & 53.9
& 200 & 61.4 & 49.5 & 37.1 & 52.4 
& 30 & 40.6
& 200 & 70.9 & A \\

\multicolumn{1}{l|}{SuperDisco \cite{SuperDisco}}& 2023 & \it CVPR 
& 200 & 53.8 & 58.3
& 100 & 66.1 & 53.3 & 37.1 & 57.1 
& 30 & 40.3
& 100 & 73.6 & A \\

\multicolumn{1}{l|}{GLMC \cite{GLMC}}& 2023 & \it CVPR 
& 200 & 57.1 & \ubold{\underline{62.3}}
& - & - & - & - & - 
& 30 & -
& 100 & - & A \\

\multicolumn{1}{l|}{BalPoE \cite{BalPoE}}& 2023 & \it CVPR 
& 200 & 52.0 & 56.3
& 180 & 66.0 & 56.7 & \ubold{\underline{43.6}} & 58.5 
& 30 & -
& 100 & 75.0 & M \\

\multicolumn{1}{l|}{SHIKE \cite{SHIKE}   }& 2023 & \it CVPR 
& 200 & 56.3 & 59.8
& 220 & - & - & - & 59.7 
& 30 & 41.9
& 220 & 75.4 & M \\

\multicolumn{1}{l|}{MDCS \cite{MDCS}  }& 2023 & \it ICCV 
& 400 & 56.1 & 60.1
& 400 & - & - & - & \ubold{\underline{60.7}} 
& 30 & \ubold{\underline{42.4}}
& 400 & 75.6 & M \\

\multicolumn{1}{l|}{LGLA \cite{LGLA} } & 2023 & \it ICCV 
& 400 & \ubold{\underline{57.2}} & 61.6
& 180 & - & - & - & 59.7 
& 30 & 42.0
& 400 & \ubold{\underline{76.2}} & M \\

\multicolumn{1}{l|}{DODA(+RIDE) \cite{DODA}}& 2024 & \it ICLR 
& 200 & 50.2 & 52.6
& 100 &66.6 & 51.9 & 35.9 & 55.8
& - & -
& 100 & 73.7 & D \\

\hhline{================}

\multicolumn{3}{l}{\ubold{\textit{Transformer Backbone (ViT)}}}
&\multicolumn{3}{c}{\ubold{ViT-B}} &
\multicolumn{5}{c}{\ubold{ViT-B}}
& \multicolumn{2}{c}{\ubold{ViT-B}}
& \multicolumn{2}{c}{\ubold{ViT-B}}
\\
\hline

\multicolumn{1}{l|}{VL-LTR \cite{VL-LTR}}& 2022 & \it ECCV 
& - & - & -
& - &  \ubold{\underline{84.5}} &  \ubold{\underline{74.6}} & 59.3 & 77.2
& - & 50.1 & -
& 76.8 & M \\

\multicolumn{1}{l|}{RAC \cite{RAC}}& 2022 & \it CVPR 
& - & - & -
& - & - & - & - & -
& - & 47.2
& - & 80.2 & M \\

\multicolumn{1}{l|}{LPT \cite{LPT}}& 2022 & \it ICLR 
& - &  \ubold{\underline{89.1}} & -
& - & - & - & - & -
& - & 50.1
& - & 76.1 & M \\

\multicolumn{1}{l|}{LTGC \cite{LTGC}}& 2024 & \it CVPR 
& - & - & -
& - & - & - &  \ubold{\underline{70.5}} &  \ubold{\underline{80.6}}
& - &  \ubold{\underline{54.1}}
& - &  \ubold{\underline{82.5}} & M \\

\bottomrule
\end{tabular}}
\end{table*}


\begin{table*}[!t]
\centering
\caption{The experimental settings of representative long-tailed recognition algorithms on the iNaturalist 2018 dataset, including the training settings and training strategies/tricks, i.e., bells and whistles. ``lr'', ``lr'' scheduler and ``lr decay'' denote the learning rate and the way of adjusting it downward after certain epochs as well as the learning rate decay hyper-parameter, respectively. ``weight decay'' represents the regularization / weight decay strategy to make the weights smaller, where L2 regularization is commonly adopted.  ``basic'' denotes methods that adopt basic transformation based augmentation (such as rotation, resize), while  RandAugment and CutMix denote methods that adopt more advanced data augmentation techniques for improving the LTL performance. ``BS'' and LDAM are short for Balanced Softmax Loss \cite{BALMS} and  label-distribution-aware margin loss \cite{marginimloss}, respectively. The optimizer used for all methods is SGD, with the momentum factor set to 0.9. }
\label{tab:expsetting}
\setlength{\tabcolsep}{0.8mm}
\renewcommand{\arraystretch}{1.5}{
\begin{tabular}{c|c|c|ccccc|cccccc}
\bottomrule
\multirow{3}{*}{\moren{\ubold{Method}}} & \multirow{3}{*}{\moren{\ubold{Year}}} & \multirow{3}{*}{\moren{\ubold{Venue}}} &
\multicolumn{5}{c}{\ubold{Training Settings (Hyper-Parameters)}}&\multicolumn{6}{|c}{\ubold{Bells and Whistles}}
\\

\cline{4-14}
& & & {\ubold{epoch}} & \makecell{\ubold{batch}  \\ \ubold{size}}&  {\ubold{lr}} & \makecell{\ubold{lr}  \\ \ubold{scheduler}} & \makecell{\ubold{lr}  \\ \ubold{decay}} & \makecell{\ubold{weight}  \\ \ubold{decay}}  & \makecell{\ubold{knowledge}  \\ \ubold{distillation}} & \makecell{\ubold{pre-}  \\ \ubold{training}} & \makecell{\ubold{data}  \\ \ubold{augmentation}} &\makecell{\ubold{extra}  \\ \ubold{loss}} & {\ubold{MoE}} \\ \hline

\multicolumn{1}{l|}{MetaSAug \cite{MetaSAug}} & 2021 & \it CVPR 
& 100 & 64 & 0.01 & linear &  $5 \times 10^{-4}$ &L2
 & - & - & basic & -& - \\

\multicolumn{1}{l|}{PaCo \cite{PaCo}}& 2021 & \it ICCV 
& 400 & 128 & 0.02 & cosine & $1 \times 10^{-4}$ &L2
 & - & MoCo & RandAugment & - & -\\

\multicolumn{1}{l|}{RIDE \cite{RIDE}}& 2021 & \it ICLR 
& 100 & 512 & 0.2 & linear &$2 \times 10^{-4}$  &L2
 & - & - & basic & LDAM & yes\\

\multicolumn{1}{l|}{GCL \cite{GCL}}& 2022 & \it CVPR 
& 100 & 512 & 0.1 & cosine &$1 \times 10^{-4}$  &L2
 & - & - & Mixup & - & -\\

\multicolumn{1}{l|}{BCL \cite{BCL}}& 2022 & \it CVPR 
& 100 & 256 & 0.2 & cosine & $1 \times 10^{-4}$  &L2
 & - & SimCLR & RandAugment & - & yes \\

\multicolumn{1}{l|}{SADE \cite{SADE}}& 2022 & \it NeurIPS 
& 200 & 512 & 0.2 & linear & $2 \times 10^{-4}$  &L2
 & - & - & basic & BS & yes\\

\multicolumn{1}{l|}{CMO \cite{CMO}}& 2022 & \it CVPR 
& 200 & 128 & 0.1 & linear &$2 \times 10^{-4}$  &L2
 & - & - & CutMix & LDAM & -\\

\multicolumn{1}{l|}{CUDA \cite{CUDA}}& 2023 & \it ICLR 
& 100 & 128 & 0.1 & linear &$2 \times 10^{-4}$  &L2
 & - & - & basic & LDAM & -\\

\multicolumn{1}{l|}{CC-SAM \cite{CC-SAM}}& 2023 & \it CVPR 
& 200 & 256 & 0.1 & linear & $2 \times 10^{-4}$  &L2
 & - & - & basic & BS & -\\

\multicolumn{1}{l|}{SHIKE \cite{SHIKE}}& 2023 & \it CVPR 
& 200+20 & 128 & 0.025 & linear & $5 \times 10^{-4}$  &L2
 & logit-distill & - & RandAugment & BS & yes\\

\multicolumn{1}{l|}{MDCS \cite{MDCS}}& 2023 & \it ICCV 
& 400 & 512 & 0.2 & linear & $5 \times 10^{-4}$ &L2
& self-distill & - & RandAugment & BS & yes\\

\multicolumn{1}{l|}{LGLA \cite{LGLA}}& 2023 & \it ICCV 
& 400 & 512 & 0.2 & cosine & $2 \times 10^{-4}$  &L2
 & - & - & RandAugment & - & yes\\

  \multicolumn{1}{l|}{DODA(+RIDE) \cite{DODA}}& 2024 & \it ICLR 
& 100 & 512 & 0.1 & linear & $2 \times 10^{-4}$  &L2
 & - & - & basic & LDAM & yes\\

 \multicolumn{1}{l|}{LTGC \cite{LTGC}}& 2024 & \it CVPR 
& - & - & - & - & -  &-
 & - & \makecell{CLIP  \\ (ViT-B)}  & \makecell{GPT-4V \& DALL-E \\Mixup}  & - & -\\
 
\bottomrule
\end{tabular}}
\end{table*}

\begin{table}[!t]

\centering
\caption{Performance of representative long-tailed object detection and segmentation   methods on  LVIS v1.0.}
\label{tab:seg_dec}
\setlength{\tabcolsep}{1.1mm}
\renewcommand{\arraystretch}{1.5}{
\begin{tabular}{c|c|c|cccc|c}
\bottomrule
\multirow{2}{*}{\moren{\ubold{Method}}} & \multirow{2}{*}{\moren{\ubold{Year}}} & \multirow{2}{*}{\moren{\ubold{Venue}}} &
\multicolumn{4}{c}{\ubold{Seg.}}&\multicolumn{1}{|c}{\ubold{Det.}}
\\
\cline{4-8}
& & & \boldmath{{${AP}$}} & \boldmath{{${AP}_{r}$}} & \boldmath{{${AP}_{c}$}} & \boldmath{{${AP}_{f}$}} & \boldmath{{${AP}^{b}$}} \\ \hline

\multicolumn{1}{l|}{BAGS \cite{li2020overcoming}}& 2020 & \it CVPR 
& 23.1 & 13.1 & 22.5 & 28.2 & 25.8\\

\hline

\multicolumn{1}{l|}{FASA \cite{FASA}}& 2021 & \it ICCV 
& 24.1 & 17.3 & 22.9 & 28.5 & -\\

\multicolumn{1}{l|}{RIO \cite{RIO}}& 2021 & \it ICML 
& 23.7 & 15.2 & 22.5 & 28.8 & 24.1\\

\multicolumn{1}{l|}{DisAlign \cite{DisAlign}}& 2021 & \it CVPR 
& 24.3 & 8.5 & 26.3 & 28.1 & 23.9\\

\multicolumn{1}{l|}{MOSAICOS \cite{MosaicOSLT}}& 2021 & \it ICCV 
& 24.5 & 18.2 & 23.0 & 28.8 & 25.0\\

\multicolumn{1}{l|}{RS \cite{RS-Loss}}& 2021 & \it ICCV 
& 25.2 & 16.8 & 24.3 & 29.9 & 25.9\\

\multicolumn{1}{l|}{EQLv2\cite{equalossv2}}& 2021 & \it CVPR 
& 25.5 & 17.7 & 24.3 & 29.9 & 25.9\\

\multicolumn{1}{l|}{Seesaw \cite{SeeSaw}}& 2021 & \it CVPR 
& 26.4 & 19.6 & 26.1 & 29.8 & 27.4\\

\multicolumn{1}{l|}{LOCE \cite{EquiLTOD}}& 2021 & \it ICCV 
& 26.6 & 18.5 & 26.2 & 30.7 & 27.4\\

\multicolumn{1}{l|}{NORCAL \cite{NORCAL}}& 2021 & \it NeurIPS 
& 26.8 & 23.9 & 25.8 & 29.1 & 27.8\\

\hline

\multicolumn{1}{l|}{FreeSeg \cite{FreeSeg}}& 2022 & \it ECCV 
& 25.2 & 20.2 & 23.8 & 28.9 & 26.0\\

\multicolumn{1}{l|}{AHRL \cite{AHR}}& 2022 & \it CVPR 
& 25.7 & - & - & - & 26.4 \\

\multicolumn{1}{l|}{C2AM \cite{C2AM}}& 2022 & \it CVPR 
& 27.2 & 16.6 & 27.2 & 31.9 & 27.9 \\

\multicolumn{1}{l|}{PCB \cite{PCB}}& 2022 & \it CVPR 
& 27.2 & 19.0 & 27.1 & 30.9 & 28.1\\

\multicolumn{1}{l|}{ECM \cite{ECM}}& 2022 & \it ECCV 
& 27.4 & 19.7 & 27.0 & 31.1 & 27.9 \\

\multicolumn{1}{l|}{GOL \cite{GOL}}& 2022 & \it ECCV 
& 27.7 & 21.4 & 27.7 & 30.4 & 27.5\\

\hline

\multicolumn{1}{l|}{Step-wise \cite{Dong_2023_ICCV}}& 2023 & \it ICCV 
& - & - & - & - & 28.7\\

\multicolumn{1}{l|}{RichSem \cite{RichSem}}& 2023 & \it NIPS 
& - & - & - & - & 30.6\\

\hline

\multicolumn{1}{l|}{BSGAL \cite{BSGAL}}& 2024 & \it ICML 
& \ubold{\underline{31.6}} & \ubold{\underline{25.4}} & \ubold{\underline{30.6}} & \ubold{\underline{35.4}} & \ubold{\underline{35.4}}\\

\bottomrule
\end{tabular}}
\end{table}

\section{Future Directions}  \label{sec:future}
While numerous approaches have been proposed to address the long tail issue, many research opportunities remain.

\subsection{Federated Long-Tailed Learning}
Federated learning enables multiple clients to learn a global model collaboratively without transmitting the local private data on each client to a centralized server \cite{CReFF}. The problem becomes more challenging when the training data across the clients is both heterogeneous and long-tailed. 

Fairness refers to whether a model can perform equally for diverse types of categories. Although there are a few preliminary studies \cite{FedFV}, ensuring fairness within a federated learning system in long-tailed settings still needs in-depth investigations.

\subsection{Long-Tailed Out-of-Distribution Detection}

Out-of-Distribution (OOD) samples are samples that do not match the training data distribution, regardless of whether they belong to known (existing) categories or entirely new (open-set) categories. In \cite{PASCL}, the authors point that it is sometimes challenging to distinguish between the in-distribution tail class samples and the OOD samples that belong to entirely new (open-set) categories. To address this issue, \cite{EAT} expands the in-distribution class space by introducing multiple abstention classes and adopts mixture of experts for classification. It also augments the context-limited tail classes using both the head and OOD classes via Cutmix. At test time, the sum of probabilities for the k abstention classes is used for OOD detection. In \cite{openlongtail}, the authors also investigate the open-set long-tailed recognition problem, in which a long-tailed distribution also contains samples of new/unknown/open-set categories. They propose to simultaneously handle long-tailed classification, few-shot learning and open-set recognition (OOD detection) in a unified framework. 

\subsection{Active Learning for Long-Tailed Distributions}

Active learning approaches iteratively select the most informative samples from underrepresented classes for annotation and training to optimize the use of limited labeled data to improve model performance over time. The authors in \cite{VaB-AL}  incorporate class imbalance into active learning by considering both class imbalance and labelling cost. \cite{PAL} proposes a progressive sampling mechanism that explores active learning of OOD samples for enhancing the performance in both long-tailed classification and OOD detection. 

\subsection{Long-Tail Class Incremental Learning}

To acquire novel knowledge from long-tailed class-incremental settings, \cite{LT-CIL} proposes to use sub-protypes to discover and store data distributions for sub-tasks with long-tailed distributions, and adopts the knowledge distillation technique to avoid catastrophic forgetting when training a new task. 

\subsection{Long-tailed  Domain Generalization}

Domain generalization aims to learn from multiple training domains to generalize to previously unseen target domains. In \cite{LTDG}, the authors introduces the maximum square loss which has a linearly increasing gradient to address the long-tailed distribution issue in domain generalization.  While \cite{LTDG} focuses on single domain imbalance, \cite{MDLT} investigates  the multiple domains LTL problem which aims to learn from multi-domain imbalanced data and generalize to all domains. It proposes a loss for bounding the transferability statistics based on the contrastive loss over the positive and negative cross-domain pairs. 

\subsection{Adversarial Training Under Long-Tailed Distributions}

Adversarial training is a defense strategy aimed at enhancing the robustness of machine learning models against adversarial attacks by incorporating adversarial examples into the training process. 
\cite{LTAT} investigates the adversarial vulnerability in long-tailed recognition and proposes RoBal, which includes a scale-invariant cosine classifier and a two-stage LTL framework. In \cite{AT-BSL}, the authors reveal that adversarial training under long-tailed distributions suffers from robust overfitting, i.e., the long-tailed model overfits to current adversarial samples. They show that simply combining Balanced Softmax Loss \cite{BALMS} and data augmentation can simultaneously alleviate robust overfitting and improve model robustness against adversarial attacks. 

\subsection{Diverse Applications with Long Tail Concerns}
Long-tailed data widely exists in many real-world domains, e.g., defect identification \cite{LTAD},  remote sensing \cite{rsLT}, nutrition ingredients analysis \cite{foodLT}, etc. While we can apply long-tailed learning techniques in these domains, such real-world applications often pose new challenges to be addressed,  which  will advance the long-tailed learning techniques in turn. 

\section{Conclusions}  \label{sec:conclusion}
In this paper, we propose a new taxonomy for long-tailed learning that divides existing techniques into eight categories, which are data balancing, neural architecture, feature enrichment, logits adjustment, loss function, bells and whistles, network optimization, and post hoc processing techniques. Based on this taxonomy, we systematically review long-tailed learning methods in each category, including the latest advances. We also examine the distinctions between imbalance learning and long-tailed learning approaches. We summarize the experimental results of representative methods, and show that there is still a very large room for further improvements. Finally, we discuss future directions in this field.

\footnotesize
\bibliographystyle{IEEEtran}
\bibliography{LT.bib}

\end{document}